\documentclass[10pt,twocolumn,letterpaper]{article}

\usepackage[pagenumbers]{cvpr} 
\usepackage[font=footnotesize,labelfont=bf]{caption} 
\usepackage[accsupp]{axessibility}  
\usepackage[pagebackref,breaklinks,colorlinks,citecolor=cvprblue]{hyperref}
\usepackage[capitalize]{cleveref}
\usepackage[dvipsnames]{xcolor}

\newcommand{\tickmark}{\color{teal}\bf{\checkmark}}
\newcommand{\xmark}{\color{red}$\bf{\times}$}
\newcommand{\adashell}{AdaShell$^{++}$}
\newcommand{\volsdf}{VolSDF$^{++}$}
\newcommand{\neus}{NeUS$^{++}$}
\newcommand{\unisurf}{UniSurf$^{++}$}
\definecolor{ForestGreen}{RGB}{34,139,34}

\definecolor{cvprblue}{rgb}{0.21,0.49,0.74}

\def\paperID{357} 
\def\confName{3DV\xspace}
\def\confYear{2025\xspace}

\title{Incorporating dense metric depth into neural 3D representations for view synthesis and relighting}

\author{Arkadeep Narayan Chaudhury\\
Carnegie Mellon Uiversity\\
\and
Igor Vasiljevic\\
Toyota Research Institute\\
\and
Sergey Zakharov\\
Toyota Research Institute\\
\and
Vitor Guizilini\\
Toyota Research Institute\\
\and
Rares Ambrus\\
Toyota Research Institute\\
\and
Srinivasa Narasimhan\\
Carnegie Mellon University\\
\and
Christopher G. Atkeson\\
Carnegie Mellon University\\
}
\newcommand{\shrink}{\def\baselinestretch{0.95}\large\normalsize} 
\shrink

\begin{document}
\twocolumn[{%
\renewcommand\twocolumn[1][]{#1}%
\maketitle
\centering
\includegraphics[width=\textwidth]{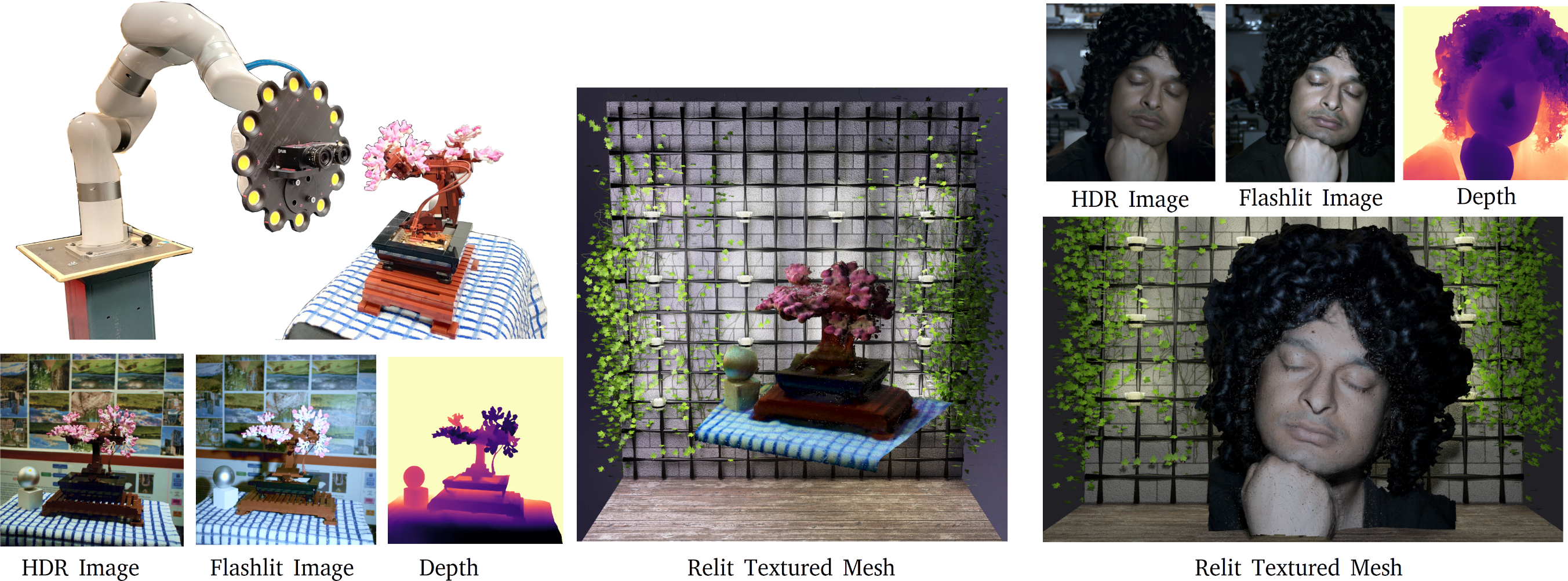}
\captionof{figure}{We present an approach for the photo-realistic capture of small scenes by incorporating dense metric depth, multi-view, and multi-illumination images into neural 3D scene understanding pipelines. We use a robot mounted multi-flash stereo camera system, developed in-house, to capture the necessary supervision signals needed to optimize our representation with a few input views. The reconstruction of the LEGO plant and the face were generated with 11 and 2 stereo pairs respectively. We relight the textured meshes using \cite{blender}. Background design by \cite{Ginibird}.}
        \vspace{.15in}
\label{fig:teaser}
}]
\maketitle
\begin{abstract}
Synthesizing accurate geometry and photo-realistic appearance of small scenes is an active area of research with compelling use cases in gaming, virtual reality, robotic-manipulation, autonomous driving, convenient product capture, and consumer-level photography. When applying scene geometry and appearance estimation techniques to robotics, we found that the narrow cone of possible viewpoints due to the limited range of robot motion and scene clutter caused current estimation techniques to produce poor quality estimates or even fail. On the other hand, in robotic applications, dense metric depth can often be measured directly using stereo and illumination can be controlled. Depth can provide a good initial estimate of the object geometry to improve reconstruction, while multi-illumination images can facilitate relighting. In this work we demonstrate a method to incorporate dense metric depth into the training of neural 3D representations and address an artifact observed while jointly refining geometry and appearance by disambiguating between texture and geometry edges. We also discuss a multi-flash stereo camera system developed to capture the necessary data for our pipeline and show results on relighting and view synthesis with a few training views. 
\end{abstract}

\section{Introduction}
\label{sec:intro}
Capturing photo realistic appearance and geometry of scenes is a fundamental problem in computer vision and graphics with a set of mature tools and solutions for content creation \cite{Hyper-capture, Luma_AI}, large scale scene mapping \cite{3DZephyr}, augmented reality and cinematography \cite{RealityCapture,AliceVision,nerfstudio}. 
Enthusiast level 3D photogrammetry, especially for small or tabletop scenes, has been supercharged by more capable smartphone cameras and new toolboxes like RealityCapture and NeRFStudio. A subset of these solutions are geared towards view synthesis where the focus is on photo-realistic view interpolation rather than recovery of accurate scene geometry. These solutions take the ``shape-radiance ambiguity''\cite{kutulakos1999theory} into stride by decoupling the scene transmissivity (related to geometry) from the scene appearance prediction. But without of diverse training views, several neural scene representations (e.g. \cite{mildenhall2021nerf, muller2022instantngp, fridovich2022plenoxels}) are prone to poor shape reconstructions while estimating accurate appearance. 
\newline \indent By only reasoning about appearance as cumulative radiance weighted with the scene's transmissivity, one can achieve convincing view interpolation results, with the quality of estimated scene geometry improving with the diversity and number of training views. However, capturing a diverse set of views, especially for small scenes, often becomes challenging due to the scenes' arrangement. 
\newline \indent Without of dense metric depth measurements, researchers have used sparse depth from structure from motion \cite{gaussian_splatting,kangle2021dsnerf,roessle2022depthpriorsnerf}, and dense monocular depth priors \cite{Yu2022MonoSDF} to improve reconstruction, with a focus on appearance. Assimilating dense non-metric depth (e.g.  \cite{eftekhar2021omnidata,cheng2019convspatialprop}) is often challenging due to the presence of an unknown affine degree of freedom which needs to be estimated across many views.     
\newline \indent However, without diversity of viewpoints, measuring the geometry directly is often useful. Several hardware solutions for digitizing objects exist, ranging from consumer level 3D scanners (e.g. \cite{Einscan_2023}), and room scale metrology devices (\cite{Ensenso_2023, Photoneo_2023, Matterport}) to high precision hand held 3D scanners (e.g. \cite{Artec3D_2023}). Although these systems measure geometry very accurately, they interpret appearance as diffuse reflectance and often fall short in modelling view-dependent effects.
\newline \indent Despite the known effectiveness of incorporating depth and widespread availability of dense metric depth sensors in smartphone cameras (\cite{zhangDu2Net20202, ZDNet}) and as standalone devices (\cite{Keselman2017_realsense,azure_kinect_2023}, incorporation of dense metric depth into neural 3D scene understanding is underexplored. 
In this work:
\begin{enumerate}
\item we present a method to incorporate dense metric depth into the training of neural 3D fields, enabling state-of-the-art methods to use dense metric depth with minor changes. 
\item We investigate an artifact (\cref{fig:unisurf_pathology}) commonly observed while jointly refining shape and appearance. We identify its cause as existing methods' inability to differentiate between depth and texture discontinuities. We address it by using depth edges as an additional supervision signal. 
\end{enumerate}
We demonstrate our ideas using a robot mounted multi-flash stereo camera rig developed in-house from off-the-shelf components. This device allows us to capture a diverse range of scenes with varying complexity in both appearance and geometry. Using the captured data, we demonstrate results in reconstruction, view interpolation, geometry capture, and relighting with a few views. We hope that our full-stack solution comprising of the camera system and algorithms will serve as a test bench for automatically capturing small scenes in the future. Additional results may be viewed at \href{https://stereomfc.github.io}{https://stereomfc.github.io} and in the supplementary document. 
\begin{figure*}
\centering
\begin{subfigure}[b]{0.18\textwidth}
\centering
\includegraphics[width=0.90\textwidth]{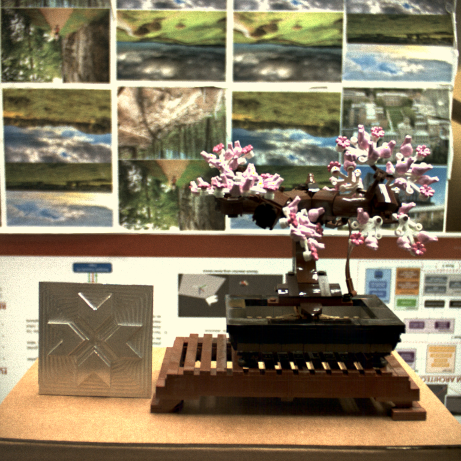}
\caption{}
\label{fig:mfc_main_RGB}
\end{subfigure}
\begin{subfigure}[b]{0.21\textwidth}
\centering
\includegraphics[width=0.92\textwidth]{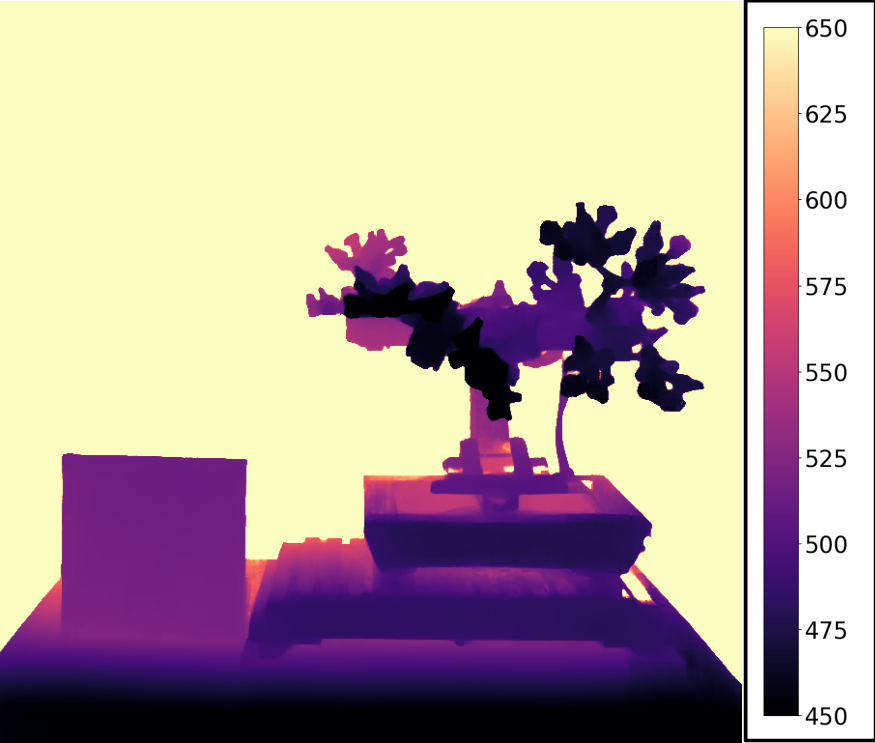}
\caption{}
\label{fig:mfc_stereo_depth}
\end{subfigure}
\begin{subfigure}[b]{0.18\textwidth}
\centering
\includegraphics[width=0.90\textwidth]{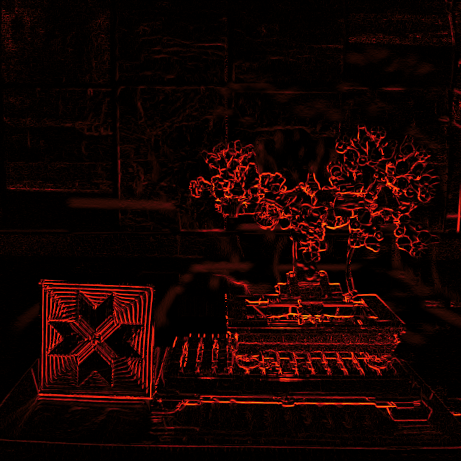}
\caption{}
\label{fig:mfc_depth_edges}
\end{subfigure} 
\begin{subfigure}[b]{0.18\textwidth}
\centering
\includegraphics[width=0.90\textwidth]{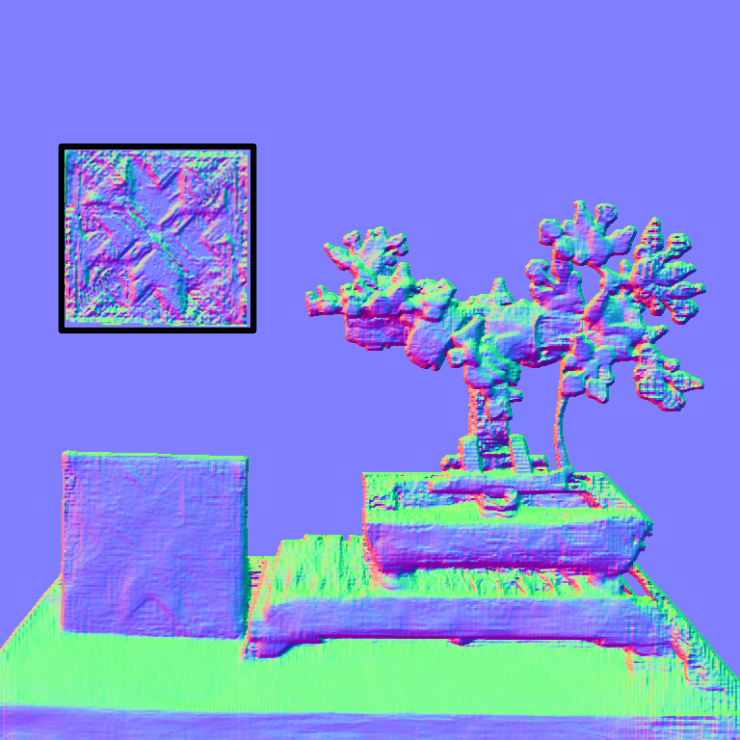} 
\caption{}
\label{fig:mfc_normals}
\end{subfigure}
\begin{subfigure}[b]{0.18\textwidth}
\centering
\includegraphics[width=0.90\textwidth]{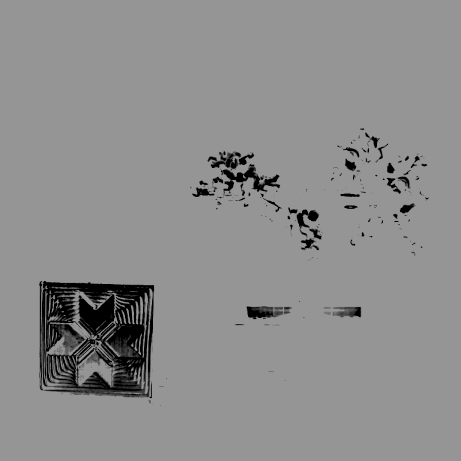} 
\caption{}
\label{fig:mfc_specularity_labels}
\end{subfigure}
\caption{\textbf{A snapshot of the important supervision signals.} We capture a high dynamic range image \cite{mertens2007exposure} and display it after tonemapping \cite{reinhard2002} in \cref{fig:mfc_main_RGB}. \Cref{fig:mfc_stereo_depth} shows the scene depth (in mm) from stereo. \Cref{fig:mfc_depth_edges} displays the likelihood of each pixel falling on a depth edge. \Cref{fig:mfc_normals} shows the object surface normals. We note that unlike conventional stereo matching (\cite{hirschmuller2005accurate}), \cite{xu2022gmflow} returns locally smooth surfaces and often ignores local texture variations but is less noisy. The inset shows the surface normals on the textured aluminum plate calculated as gradients of depth from conventional stereo matching. Finally, \cref{fig:mfc_specularity_labels} identifies the pixels with the largest appearance variation due to moving lights. We used the system in \cref{fig:MFC_schematic} to capture the data.}
\label{fig:mfc_main}
\end{figure*}
\section{Related Work}\label{sc:related_work}
\textbf{View synthesis} and reconstruction of shapes from multiple 3D measurements is an important problem in computer vision with highly efficient and general solutions like volumetric fusion (\cite{curless1996volumetric}), screened Poisson surface reconstruction (\cite{kazhdan2013screened}), patch based dense stereopsis (\cite{Galliani_2015_ICCV_GIPUMA}) and joint refinement of surface and appearance \cite{dai2017bundlefusion}. While these continue to serve as robust foundations, they fall short in capturing view-dependent appearance. Additionally, even with arbitrary levels of discretization, they often oversmooth texture and surfaces due to data association relying on weighted averages along the object surface. 
\newline
\indent Recent neural 3D scene understanding approaches (e.g. \cite{yariv2021volsdf,wang2021neus,li2023neuralangelo}) have avoided this by adopting a continuous implicit volumetric representation to serve as the geometric and appearance back-end of the view synthesizer. Together with continuous models, reasoning about appearance as radiance, and high frequency preserving embeddings\cite{tancik2020fourier}, these approaches serve as highly capable view interpolators by reliably preserving view dependent appearance and minute geometric details. More recent work has included additional geometric priors in the form of monocular depth supervision (\cite{Yu2022MonoSDF}), sparse depth supervision from structure-from-motion toolboxes (\cite{sun2022neuralreconW}), dense depth maps (\cite{Azinovic_2022_NeuralRGBD,Sandstrom2023ICCV}), patch based multi-view consistency \cite{fu2022geoneus}, and multi-view photometric consistency under assumed surface reflectance functions (\cite{guizilini2023delira}). Our work builds on the insights from using dense depth supervision to improve scene understanding with only a few training views available. 
\newline
\indent \textbf{Novel hardware} is often used for collecting supervision signals in addition to color images to aid 3D scene understanding. \cite{attal2021torf} demonstrate a method to incorporate a time-of-flight sensor. \cite{Shandilya_2023_ICCV} demonstrate a method to extract geometric and radiometric cues from scenes captured with a commercial RGBD sensor and improve view synthesis with a few views. Event based sensors have also been used to understand poorly lit scenes with fast moving cameras (\cite{klenk2023ev-nerf,Low_2023_ICCV-ev-nerf}). Researchers have also combined illumination sources with cameras to capture photometric and geometric cues for dense 3D reconstruction of scenes with known reflectances (\cite{Gotardo_2015_ICCV,chaudhury2024shape}). Similarly, \cite{schmitt2023towards,Schmitt_2020_CVPR,cheng2023wildlight} capture geometry and reflectance of objects by refining multi-view color, depth and multi-illumination images. Given the recent advances in stereo matching (\cite{xu2022gmflow}) we use a stereo camera to collect data for view synthesis to disambiguate between shape and appearance at capture. 
\newline \indent \textbf{Pairing illumination} sources with imaging can improve reasoning about the appearance in terms of surface reflectance parameters. \cite{kaizhang_kang2019learning,schmitt2023towards,zhou2013multi} approaches the problem of material capture using a variety of neural and classical techniques. \cite{zhang2022iron,cheng2023wildlight,bi2020neural,kang2021neural} leverage recent neural scene understanding techniques to jointly learn shape and appearance as reflectance of the scene. Our work also pairs illumination sources with stereo cameras to capture multi-illumination images from the scene and we build on modern neural techniques for view synthesis and relighting the scene. 
\section{Methods}\label{sc:methods}
We follow related works (\cite{yariv2021volsdf,wang2021neus,li2023neuralangelo,adaptiveshells2023}) and represent the scene with two neural networks -- an intrinsic network $\mathcal{N}(\theta)$ and an appearance network $\mathcal{A}(\phi)$ which are jointly optimized to capture the shape and appearance of the object. $\mathcal{N}(\theta)$ is a multi-layer perceptron (MLP) with parameters $(\theta)$ and uses multi-level hash grids to  encode the inputs (\cite{li2023neuralangelo}). It is trained to approximate the intrinsic properties of the scene -- the scene geometry as a neural signed distance field $\mathcal{S}(\theta)$ and an embedding $\mathcal{E}(\theta)$. The appearance network $\mathcal{A}(\phi)$ is another MLP which takes $\mathcal{E}(\theta)$ and a frequency encoded representation of the viewing direction and returns the scene radiance along a ray. 
\newline \indent Prior work has jointly learned $\mathcal{S, N, A}$ with only multi-view images by optimizing a loss in the form of \cref{eq:total_reconstruction_loss} using stochastic gradient descent \cite{kingma2014adam} along a batch of rays projected from known camera centers to the scene. 
\begin{equation}\label{eq:total_reconstruction_loss}
    \ell = \ell_C + \lambda_{g} \ell_D + \lambda_{c} \mathbb{E}(|\nabla^2_\mathbf{x} \mathcal{S}(\mathbf{x}_s)|)
\end{equation}
$\lambda$s are hyperparameters and the third term in \cref{eq:total_reconstruction_loss} is the mean surface curvature minimized against the captured surface normals (see \cite{li2023neuralangelo}). As the gradients of the loss functions $\ell_C$ (appearance loss)  and $\ell_D$ (geometry loss) propagate through $\mathcal{A}$ and $\mathcal{N}$ (and $\mathcal{S}$ as it is part of  $\mathcal{N}$) the appearance and geometry are learned together.
\newline \indent We describe our method of incorporating dense metric depth in \cref{sc:sdf_with_depth} which enables a variety of neural 3D representations (\cref{sc:baselines}) to use it. In \cref{sc:appearance_and_shape} we jointly optimize shape and appearance of a scene using information about scene depth edges. 
\subsection{Incorporating dense metric depth}\label{sc:sdf_with_depth}
Given a large number of orthogonal view pairs (viewpoint diversity), and the absence of very strong view dependent effects, \cref{eq:total_reconstruction_loss} is expected to guide $\mathcal{S}$ to towards an unbiased estimate of the true scene depth (see e.g. \cite{BayesRaysGoli2023}). We can accelerate the convergence by providing high quality biased estimate of the scene depth. Given the quality of modern deep stereo (\cite{xu2022gmflow}) and a well calibrated camera system, a handful of aligned RGBD sequences can serve as a good initial estimate of the true surface depth in absence of diverse viewpoints. 
\newline \indent In this section we describe our method to directly optimize $\mathcal{S}$ with estimates of true surface depth to any surface point $\mathbf{x}_s$. 
Although \cite{oechsle2021unisurf, dai2017bundlefusion, zollhofer2015shading} use the depth estimates directly, they fall short of modelling view-dependent effects. To avoid that, we elect to learn a continuous and locally smooth function that approximates the signed distance function of the surface $\mathbf{x}_s$ which can then be transformed to scene density (\cite{yariv2021volsdf, wang2021neus, oechsle2021unisurf}). To do this, we roughly follow \cite{gropp2020IGR} and consider a loss function of the form 
\begin{align}\label{eq:igr_loss}
    \ell_{D}(\theta) &= \ell_{\mathbf{x}_s} + \lambda \mathbb{E}(||\nabla_{\mathbf{x}}\mathcal{S}(\mathbf{x}^{\Delta}, \theta)|| - 1)^2\\
    \text{where, } \ell_{\mathbf{x}_s} &= \frac{1}{N}\Sigma_{\forall \mathbf{x}} \left[ \mathcal{S}(\mathbf{x}, \theta) + 1-\langle \nabla_{\mathbf{x}}\mathcal{S}(\mathbf{x}, \theta), \mathbf{n}_x \rangle \right] \nonumber
\end{align}
Through the two components of $\ell_{\mathbf{x}_s}$, the loss encourages the function $\mathcal{S}(\mathbf{x}, \theta)$ to vanish at the observed surface points and the gradients of the surface to align at the measured surface normals $(\mathbf{n}_x)$. The second component in \cref{eq:igr_loss} is the Eikonal term (\cite{crandall1983viscosity}) which encourages the gradients of $\mathcal{S}$ to have a unit $L_2$ norm everywhere. The individual terms of \cref{eq:igr_loss} are averaged across all samples in a batch corresponding to $N$ rays projected from a known camera. 
\newline 
\indent The Eikonal constraint applies to the neighborhood points $\mathbf{x}^{\Delta}_s$ of each point in $\mathbf{x}_s$. \cite{gropp2020IGR} identifies candidate $\mathbf{x}^{\Delta}_s$ through a nearest neighbor search, where as \cite{yariv2021volsdf} identifies $\mathbf{x}^{\Delta}$ through random perturbations of the estimated surface point along the projected ray. As we have access to depth maps, we identify the variance of the neighborhood of $\mathbf{x}_s$ through a sliding window maximum filter on the depth images. This lets us avoid expensive nearest neighbor lookups for a batch of $\mathbf{x}_s$ to generate better estimates of $\mathbf{x}^{\Delta}_s$ than \cite{yariv2021volsdf} at train time. As a result, convergence is accelerated -- ($\sim 100\times$ over \cite{gropp2020IGR}) with no loss of accuracy. As we used metric depth, noisy depth estimates for parts of the scene are implicitly averaged by $\mathcal{S}$ optimized by minimizing \cref{eq:igr_loss}, making us more robust to errors than \cite{Yu2022MonoSDF}. We provide more details in the supplementary material.
\subsection{Incorporating depth edges in joint optimization of appearance and geometry}\label{sc:appearance_and_shape}
\begin{figure*}
\centering
\begin{subfigure}[b]{0.145\textwidth}
\centering
\includegraphics[width=0.90\textwidth]{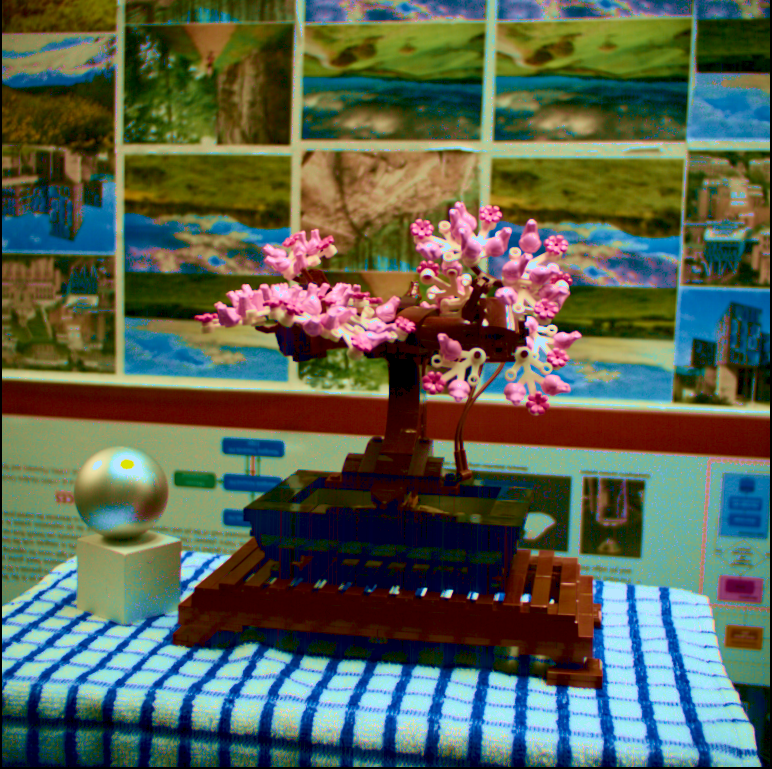}
\caption{}
\label{fig:lego_plant_gt}
\end{subfigure}
\begin{subfigure}[b]{0.145\textwidth}
\centering
\includegraphics[width=0.92\textwidth]{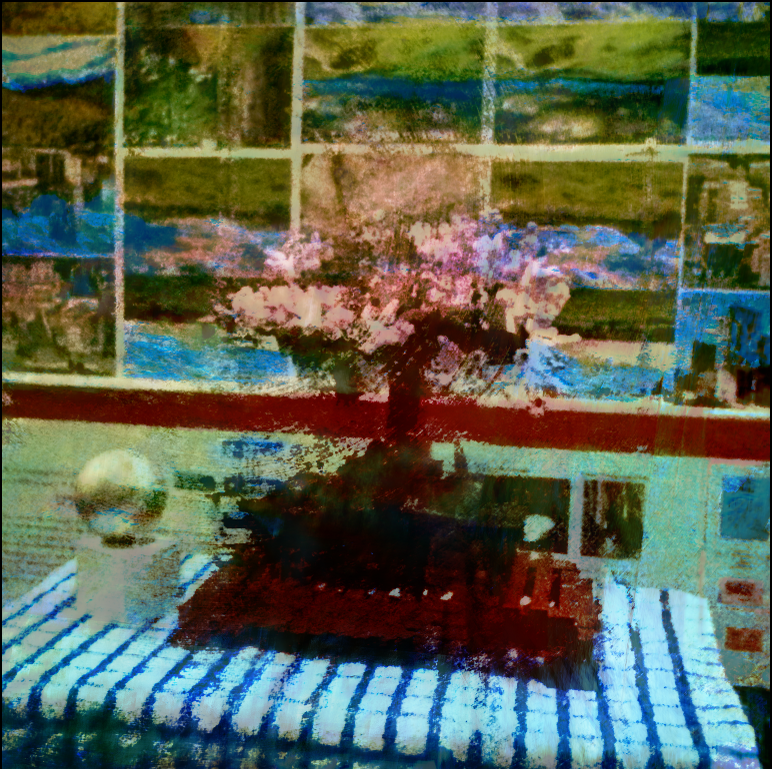}
\caption{}
\label{fig:lego_plant_10_pct}
\end{subfigure}
\begin{subfigure}[b]{0.145\textwidth}
\centering
\includegraphics[width=0.90\textwidth]{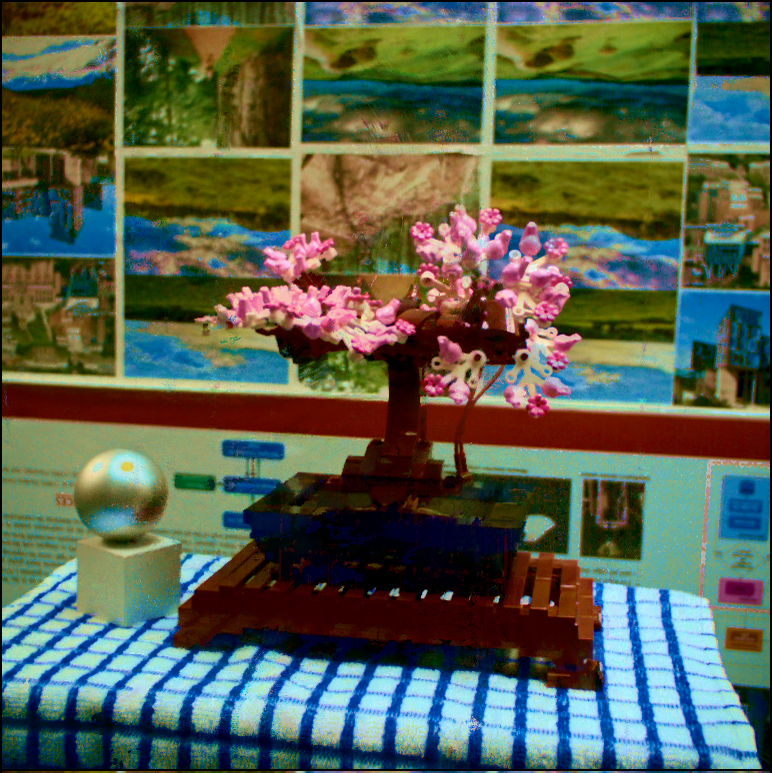}
\caption{}
\label{fig:lego_plant_100_pct}
\end{subfigure} 
\begin{subfigure}[b]{0.145\textwidth}
\centering
\includegraphics[width=0.90\textwidth]{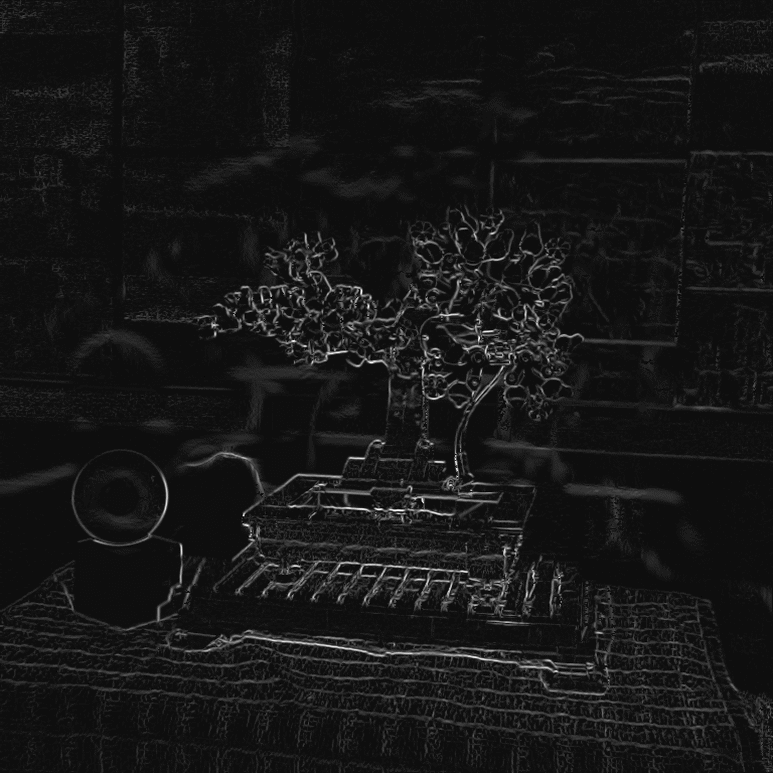} 
\caption{}
\label{fig:lego_plant_edges}
\end{subfigure}
\begin{subfigure}[b]{0.145\textwidth}
\centering
\includegraphics[width=0.90\textwidth]{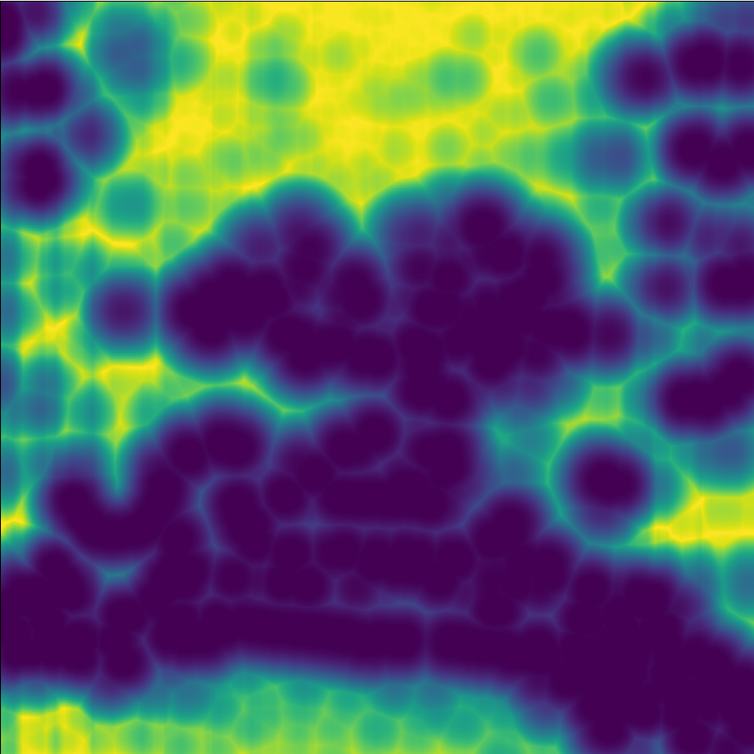} 
\caption{}
\label{fig:lego_plant_sampling_10}
\end{subfigure}
\begin{subfigure}[b]{0.145\textwidth}
\centering
\includegraphics[width=0.90\textwidth]{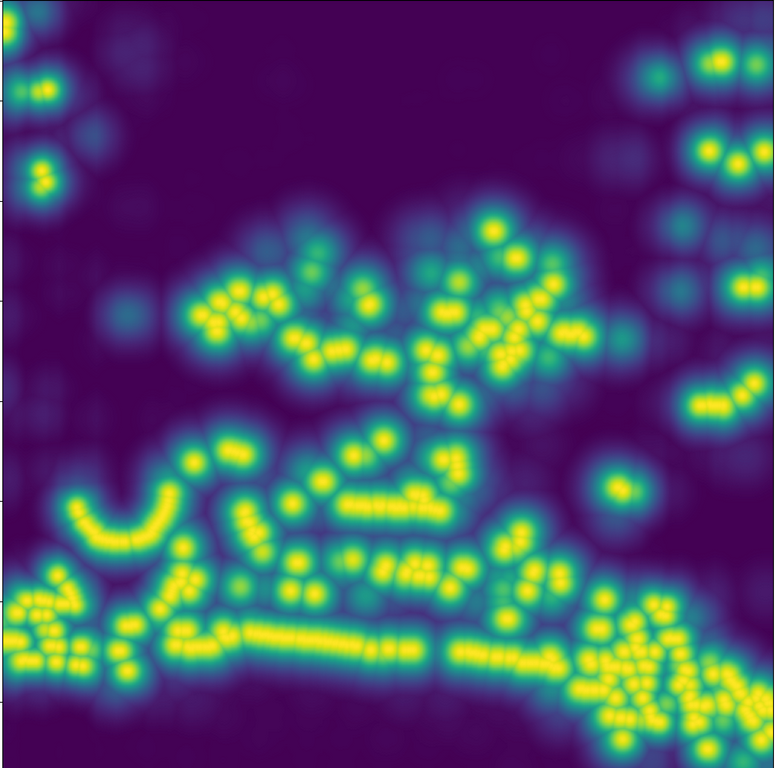} 
\caption{}
\label{fig:lego_plant_sampling_90}
\end{subfigure}
\caption{\textbf{Overview of our sampling process during training}. \Cref{fig:lego_plant_gt} is the ground truth test image. \Cref{fig:lego_plant_10_pct} is the reconstruction of the test image after training has progressed 15\% (15k gradient steps), \cref{fig:lego_plant_100_pct} is the reconstruction of the test image at the end of training (100k gradient steps). \Cref{fig:lego_plant_edges} denotes the per-pixel likelihoods of depth edges in the scene at the same view captured with our device. We note in \cref{fig:lego_plant_10_pct}, the parts of scene with complicated geometry (foliage with many depth edges) have lower fidelity of appearance in the reconstruction at an earlier stage of training, which gradually improves in \cref{fig:lego_plant_100_pct}. \Cref{fig:lego_plant_sampling_10} indicates the per-pixel sampling likelihood \textit{if the test view were to be used for training}, at a training progress of 10\%, \cref{fig:lego_plant_sampling_90} indicates the same at a progress of 90\%. \Cref{eq:edge_selection_probabililty} is used to draw the samples: $\alpha = 0.1$ and $0.9$ respectively for \cref{fig:lego_plant_sampling_90,fig:lego_plant_sampling_10}. Brighter color indicates higher sampling likelihood.}
\label{fig:sampling_main}
\end{figure*}
\begin{figure}
    \begin{subfigure}[b]{0.225\textwidth}
    \centering
    \includegraphics[width=\textwidth]{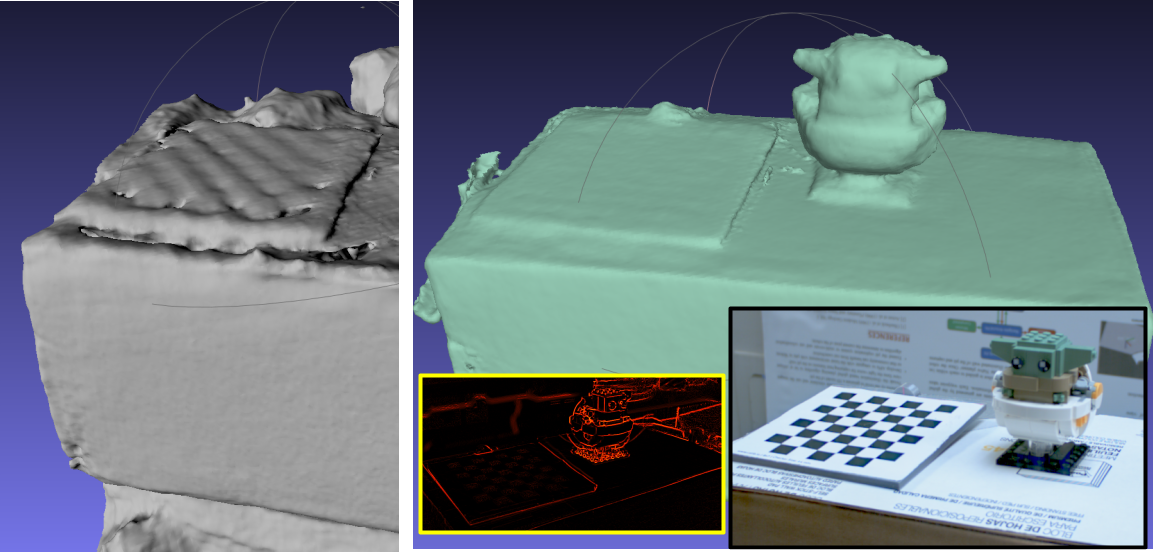}
    \caption{}
    \label{fig:unisurf_checker_horizontal}
    \end{subfigure}
    \begin{subfigure}[b]{0.24\textwidth}
    \centering
    \includegraphics[width=\textwidth]{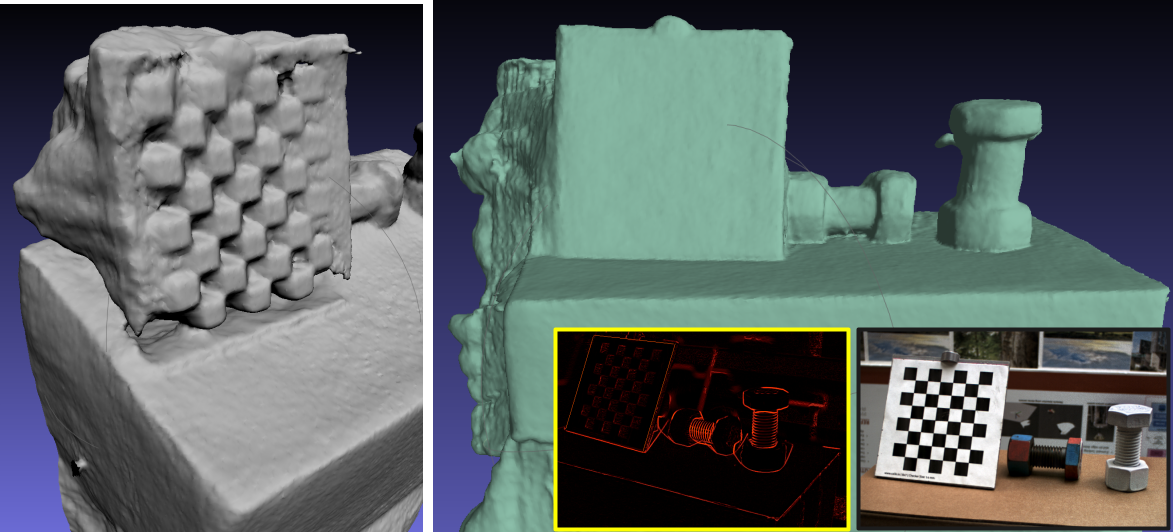}
    \caption{}
    \label{fig:unisurf_checker_vertical}
    \end{subfigure}
\caption{\textbf{We demonstrate a corner case of jointly refining appearance and geometry.} The left insets of \cref{fig:unisurf_checker_horizontal,fig:unisurf_checker_vertical} are the scene geometries recovered in the worst cases, the right insets display the better meshes recovered using the method described in \cref{sc:appearance_and_shape}. An image used for training and the edge map used for sampling are in the insets. We recommend zooming into the figure for details. Corresponding quantitative results are in \cref{tab:jointly_refining_shape_appearance}.}
\label{fig:unisurf_pathology}
\end{figure}

 
Prior works (\cite{yariv2021volsdf,brahimi2024supervol,wang2021neus,li2023neuralangelo,dai2017bundlefusion,zollhofer2015shading}) show the benefits of jointly refining geometry and appearance as it affords some degree of geometric super-resolution and more stable training. However, some pathological cases may arise when the scene has a large variation in appearance corresponding to a minimal variation in geometry across two neighboring surface points -- $\mathbf{x}_s$ and $\mathbf{x}_s^\Delta$. We investigate this effect by considering an extreme case -- a checkerboard printed on matte paper with an inkjet printer, where there is no geometric variation (planar geometry) or view dependent artifacts (ink on matte paper is close to Lambertian) corresponding to a maximum variation in appearance (white on black). The qualitative results are presented in \cref{fig:unisurf_pathology}. 
\newline \indent Consider two rays $\vec{r}_{\mathbf{x}_s}$ and $\vec{r}_{\mathbf{x}_s^\Delta}$ connecting the camera center and two neighboring points $\mathbf{x}_s$ and $\mathbf{x}_s^\Delta$ on two sides of an checkerboard edge included in the same batch of the gradient descent. The total losses for those rays depend on the sum of the geometry and appearance losses (\cref{eq:total_reconstruction_loss}). By default, the current state of the art (\cite{li2023neuralangelo,yariv2023bakedsdf,adaptiveshells2023} etc.) do not have a mechanism to disambiguate between texture and geometric edges (depth discontinuities).  
\newline \indent As seen in \cref{fig:unisurf_pathology}, given unsuitable hyperparameters, the approaches will continue to jointly update both geometry and appearance to minimize a combined loss (\cref{eq:total_reconstruction_loss}). 
This can often result in pathological reconstructions (left insets in \cref{fig:unisurf_pathology}) due to $\ell_C$ gradients dominating over $\ell_D$. By gradually increasing the modelling capacity of $\mathcal{N}$ we can somewhat avoid this artifact and force the gradient updates to focus on $\mathcal{A}$ to minimize the cumulative loss. \cite{li2023neuralangelo} recognize this and provide an excellent set of hyperparameters and training curricula to gradually increase the modelling capacity of $\mathcal{N}(\phi)$. This results in remarkable geometric reconstructions for well known datasets (\cite{jensen2014DTU,knapitsch2017tanksandtemples}). Alternatively, if we have per-pixel labels of geometric edges ($\mathbf{E}$, \cref{fig:mfc_depth_edges,fig:lego_plant_edges}), we can preferentially sample image patches with low variation of geometric features when the model capacity is lower ($\mathcal{S}(\theta)$ tends to represent smoother surfaces), and focus on image patches with geometric edges when the model capacity has increased. The modelling capacity of $\mathcal{A}(\phi)$ never changes. 
\newline \indent \Cref{fig:sampling_main} describes our sampling procedure while learning a scene with a variety of geometric and texture edges. \Cref{eq:edge_selection_probabililty} is used to draw pixel samples -- the probability of drawing pixel $p_i$ is calculated as a linear blend of the likelihood that it belongs to the set of edge pixels $\mathbf{E}$ and $\alpha$ is a scalar ($\alpha \in [0, 1]$) proportional to the progress of the training. 
\begin{equation}\label{eq:edge_selection_probabililty}
    P(p_i|\alpha) = (1-\alpha)P(p_i \in \mathbf{E}) + \alpha P(p_i \notin \mathbf{E})
\end{equation}
To preserve the geometric nature of the edges while ruling out high frequency pixel labels, we use Euclidean distance transform (\cite{felzenszwalb2012EDT}) to dilate $\mathbf{E}$ before applying \cref{eq:edge_selection_probabililty}. We provide  implementation details in the supplementary material for reproducibility. We discuss quantitative results in \cref{sc:experiment_depth_edges}.
\subsection{Baselines augmented with depth}\label{sc:baselines}
As baselines, we augment four state-of-the-art methods to incorporate metric depth: 
\newline \indent \textbf{\adashell} is our implementation of AdaptiveShells (\cite{adaptiveshells2023}) using metric depth. We retain the formulations for the scene geometry and appearance models, and adapt the formulation of the ``shells'' to use dense metric depth. Through \adashell we also demonstrate how dense metric depth can combine the advantages of volumetric and surface based representations in \cref{fig:schematic_of_adashell}.  
\newline \indent \textbf{\volsdf }is our augmented version of \cite{yariv2023bakedsdf,yariv2021volsdf}, where we use the metric depths along the rays to optimize the geometry \cref{eq:igr_loss}. All the other parts of the original approaches, including the methods for generating samples to minimize \cref{eq:total_reconstruction_loss} and scene density transforms are left unchanged. 
\newline \indent \textbf{\neus }is our augmented version of \cite{li2023neuralangelo}, where we also use \cref{eq:igr_loss} to optimize the geometry. The rest of the algorithm including the background radiance field is left intact.  
\newline \indent \textbf{\unisurf }is our deliberately hamstrung version of \cite{oechsle2021unisurf} where we force the the samples generated for the volumetric rendering step to have a very low variance around the current biased estimate of the surface. This makes the algorithm necessarily indifferent to the relative magnitudes of $\ell_D$ and $\ell_C$ in \cref{eq:total_reconstruction_loss}, and helps us exaggerate the pathological effects of not segregating texture and geometric edges. We choose to name the method \unisurf after we (and \cite{brahimi2024supervol}) observed that original method was vulnerable to this artifact under certain hyperparameter choices. \newline \indent  Across all the methods, we implement and train $\mathcal{N}(\theta)$ following \cite{li2023neuralangelo}, and all of them use the same appearance network $\mathcal{A}(\phi)$. \adashell and \unisurf require a warm start -- $\mathcal{S}$ pre-optimized for 5K gradient steps. All methods except \unisurf use the sampling strategy from \cref{sc:appearance_and_shape}. More details are in the supplementary material.
\section{Setup and Dataset}
\subsection{A multi-flash stereo camera}\label{sc:mfc_stereo}
\begin{figure} 
    \centering
    \includegraphics[width=0.40\textwidth]{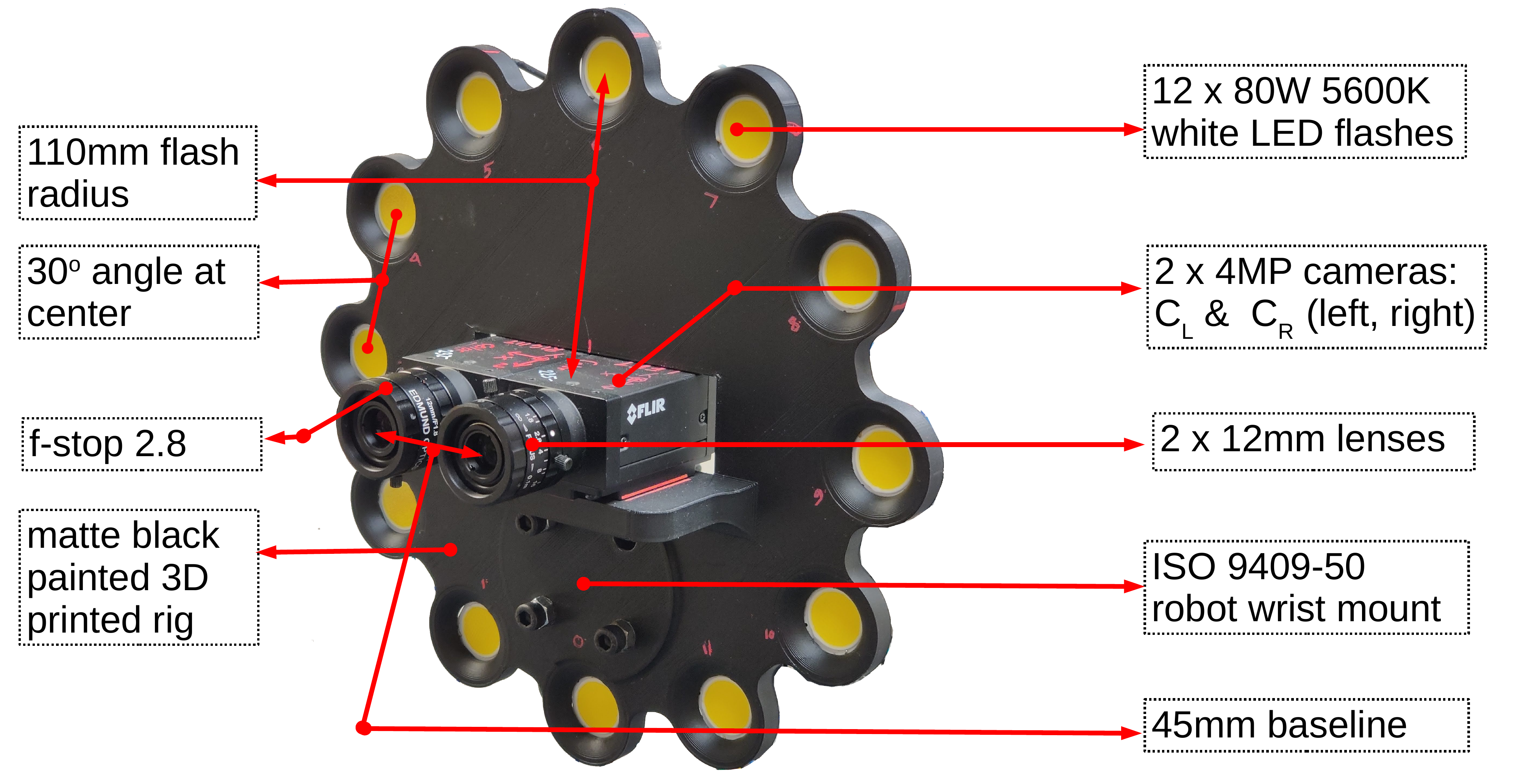}
\caption{\textbf{A multi-flash stereo camera} to image small scenes.}
\label{fig:MFC_schematic}
\end{figure}
In addition to multi-view images, scene depth and depth edges are valuable signals to train neural 3D representations. To capture all the supervision signals, we designed and fabricated a multi-flash stereo camera based on insights from \cite{raskar2004MFC,feris2004specular}, with off-the-shelf parts. We capture data by moving our camera rig in front of objects. For each camera pose we capture a stereo pair of high dynamic range (HDR) images, two depth maps from left and right stereo, two corresponding image aligned surface normals (as gradient of depth maps). We first tonemap the HDR images \cite{reinhard2002} and in-paint them with the depth edges before using \cite{xu2022gmflow} to preserve intricate surface details in the depth maps. Additionally we capture 12 pairs of multi-illumination images for 12 flash lights around the cameras, one light at a time. From the multi-flash images we recover a per pixel likelihood of depth edges in the scene and a label of pixels with a large appearance variation under changing illumination -- relating to the specularity. We detail the design of our rig and the capture processes in the supplementary material. \Cref{fig:mfc_main} shows a snapshot of the data captured, \cref{fig:MFC_schematic} illustrates our camera rig prototype. 
\newline \indent We elected to calculate depth from stereo because it performs better than the following three alternatives we tested.  1) Recovering geometry from intrinsic-image-decomposition (\cite{das2022pie}) and photometric stereo with a few lights (\cite{chaudhury2024shape}) did not yield satisfactory results. PIE-Net(\cite{das2022pie}) requires 256$\times$256 images which were too low-resolution for reconstruction and, our captures were out-of-distribution for the pre-trained model. \cite{chaudhury2024shape} assumes fixed lights -- we’d need new light calibrations per-view. 2) Self-calibrating-photometric-stereo (\cite{li2022self,chen2019self}), demonstrated on \cite{shi2016diligent}, needs ~50-80 light views and accurate masks which we do not capture. Also, our lights are much closer to the camera than \cite{li2020DiligentMV,shi2016diligent}. And, 3) modern camera-projector systems (\cite{Mirdehghan_2024_CVPR,chen_2020_autotuning}) yield better estimates of geometry than stereo, but is not fast enough to capture human subjects (\cref{fig:relighting_main}). 
\subsection{Dataset}\label{sc:dataset}
Although a data set is not the primary contribution of our research, we capture some salient aspects of the scene that are not present in several established datasets. We identify these aspects in \cref{tab:dataset_comparison}. In the rows labeled ``specularity'' and ``depth edges'' we note if the dataset has explicit labels for the specular nature of the pixel or a presence of a depth edge at that pixel respectively. Under ``illum. model'' we note if an explicit illumination model is present per scene --  we do not capture an environment illumination model, and instead provide light poses. PaNDoRa does not have explicit specularity labels but polarization measurements at pixels may be used to derive high quality specularity labels, which are better than what our system natively captures. We differentiate between ``OLAT'' (one light at a time) and ``flash'' by the location of the source of illumination. Similar to \cite{cheng2023wildlight}, our flashes are parallel to the imaging plane, located close ($\sim$0.1f) to the camera, as opposed to ReNE and OpenIllumination.
\begin{table}  
\begin{minipage}{\columnwidth}
    \centering
    \begin{footnotesize}
    \begin{tabular}{@{}l|cccccc|c@{}}
    \toprule
           &BMVS&DTU&ReNe&DGT$^{+}$&PDR&OIl.&Ours  \\
    \hline
    Depth & \tickmark & \tickmark & \xmark & \tickmark & \xmark & \xmark &\tickmark \\ 
    \hline
    Light  & OLAT & \xmark & OLAT & OLAT & \xmark & OLAT & Flash\\
    \hline 
    Pol.   & \xmark & \xmark & \xmark  & \xmark & \tickmark & \tickmark & \xmark \\
    \hline
    Spec.  & \xmark & \xmark & \xmark & \xmark & \tickmark & \xmark & \tickmark\\
    \hline
    D.E.  & \xmark & \xmark &  \xmark & \xmark & \xmark & \xmark &\tickmark \\
    \hline
    HDR  & \tickmark & \xmark & \xmark & \tickmark & \xmark & \tickmark &\tickmark \\
    \hline
    Illum. & \tickmark & \xmark & \xmark & \tickmark & \tickmark & \tickmark &\xmark\\
    \bottomrule
    \end{tabular}
\end{footnotesize}      
\end{minipage}
\caption{\textbf{We identify some differences between our dataset and a few established datasets}: BMVS\cite{yao2020blendedmvs}, DTU\cite{jensen2014DTU}, ReNe\cite{Toschi_2023_RENE}, DiLiGenT\cite{shi2016diligent}, DiLiGenT-MV\cite{shi2016diligent} (both abbreviated as DGT$^{+}$), PaNDoRa (PDR)\cite{dave2022pandora} and Open-Illumination (OIl)\cite{liu2023openillumination}. OLAT: one light at a time, Flash: camera flash $<0.1f$ away from camera, 
Pol.: polarization information, Spec.: specularity labels, D.E.: depth edge labels, HDR: High dynamic range images, Illum.: Illumination model supplied.}\label{tab:dataset_comparison}
\end{table}
\section{Experiments and results}\label{sc:results_and_apps}
\subsection{Accuracy of incorporating metric depth}\label{sc:experiment_accuracy}
We reconstruct synthetic scenes with ground truth depth from \cite{Azinovic_2022_NeuralRGBD,Sandstrom2023ICCV} to measure the accuracy of our technique. We use 12-15 RGBD images to reconstruct the scenes and train for an average of 30k gradient steps ($\sim$1500 epochs) in about 75 minutes. In contrast, \cite{Azinovic_2022_NeuralRGBD,Sandstrom2023ICCV} use  300+ RGBD tuples and 9+ hours of training on comparable hardware. Notably, \cite{Azinovic_2022_NeuralRGBD} also optimizes for noise in camera poses and reports metrics with ground truth and optimized poses. We report the best metric among these two. \cite{dai2017bundlefusion} registers the images themselves. We register the RGBD images with a combination of rigid and photometric registration (\cite{park2017coloredICP,Zhou2016FGR,sarlin2019hloc}). We present the quantitative results in \cref{tab:scene_recon_accuracy_posed,tab:scene_recon_accuracy_unposed}. We replicate or out-perform the baselines by using a fraction of the training data and gradient steps. Among all methods discussed in \cref{sc:baselines}, \adashell and \volsdf demonstrate similar performance, \neus recovers a smoother surface at the expense of  $\sim 1.25\times$ more gradient steps. Our errors on these synthetic datasets closely reflect the performance of \cite{gropp2020IGR} on approximating surfaces from low noise point clouds. These datasets do not have large view dependent appearance variations to affect the gradient updates. 
%
\begin{table}   
\begin{minipage}{\columnwidth}
    \centering
    \begin{footnotesize}
    \begin{tabular}{@{}l||cc|ccc@{}}
    \toprule
          Scene & NRGBD &  BF & \adashell  & \neus & \volsdf \\
          \hline
          \href{https://blendswap.com/blend/8381}{greenroom} & \textbf{0.013} & 0.024 & 0.015 & 0.016 &  0.014 \\ 
          \href{https://blendswap.com/blend/14449}{staircase}  & 0.045 & 0.091 & 0.024 & \textbf{0.009 } & 0.020   \\  
        \href{https://blendswap.com/blend/11801}{kitchen I}  & 0.252 & 0.234 & 0.044 & \textbf{0.036} &   0.047 \\  
        \href{https://blendswap.com/blend/5156}{kitchen II}  & \textbf{0.032} & 0.089 & 0.045 & \textbf{0.032} & 0.060   \\   
    \bottomrule
    \end{tabular}
\end{footnotesize}      
\end{minipage}
 \caption{\textbf{Accuracy of reconstruction from un-posed RGBD images.} For \emph{\textbf{un-posed}} RGBD images, we compare the accuracy of scene reconstruction using \adashell, \neus, and \volsdf  with NeuralRGBD (NRGBD) \cite{Azinovic_2022_NeuralRGBD} and BundleFusion (BF) \cite{dai2017bundlefusion}. We report normalized Chamfer distances (\underline{lower is better}) across four synthetic scenes from \cite{Azinovic_2022_NeuralRGBD}.}     
\label{tab:scene_recon_accuracy_unposed}
\end{table}
\begin{table}   
\begin{minipage}{\columnwidth}
    \centering
    \begin{footnotesize}
    \begin{tabular}{@{}l||cccccccc@{}}
    \toprule
          Scene & \texttt{R 0} &  \texttt{R 1} & \texttt{R 2} & \texttt{O 0} & \texttt{O 1} & \texttt{O 2} & \texttt{O 3} & \texttt{O 4} \\
          \hline   
          \cite{Sandstrom2023ICCV} &  0.61 & 0.41 & 0.37 & 0.38 & 0.48 & 0.54 & 0.69 & 0.72  \\  
          $AS^{++}$  & \textbf{0.11} & \textbf{0.10} & \textbf{0.09} & \textbf{0.12} & \textbf{0.06} & \textbf{0.08 }& \textbf{0.14} & \textbf{0.11}  \\ 
     \bottomrule
    \end{tabular}
\end{footnotesize}      
\end{minipage}
\caption{\textbf{Accuracy of reconstruction from posed RGBD images.} For RGBD images with \emph{\textbf{ground-truth poses}}, we compare the accuracy of reconstructing the scene between \adashell ($AS^{++}$) and PointSLAM\cite{Sandstrom2023ICCV}. We report the mean $L_1$ distances in cm (\underline{lower is better}) across eight synthetic scenes from the Replica Dataset \cite{replica19arxiv}. The scenes with prefix \texttt{R} are the room scenes, scenes with prefix \texttt{O} are the office scenes. 
}   \label{tab:scene_recon_accuracy_posed}
\end{table}
\subsection{The effect of depth edges in training}\label{sc:experiment_depth_edges}
We tested fused RGBD maps from stereo and four baselines from \cref{sc:baselines} to investigate the effect of depth and texture edges. We use edge guided sampling (\cref{sc:appearance_and_shape,eq:edge_selection_probabililty}) for all except stereo and \unisurf to prioritize learning geometric discontinuities over appearance.  We present the results in \cref{fig:unisurf_pathology,tab:jointly_refining_shape_appearance}. All of the baselines except \unisurf improve the reconstruction accuracy due to segregation of texture and depth edges. The smoothness enforced by the curvature loss in \cref{eq:total_reconstruction_loss} also improves the surface reconstruction over stereo.  
\begin{table}   
\begin{minipage}{\columnwidth}
    \centering
    \begin{footnotesize}
    \begin{tabular}{@{}l||cccc|c@{}}
    \toprule
          Scene & stereo &   \volsdf & \neus  & \adashell  & \unisurf  \\
          \hline
          \cref{fig:unisurf_checker_horizontal} & 6.22 & 4.74 & \textbf{2.77} & 5.42 & {13.21} \\ 
          \cref{fig:unisurf_checker_vertical} & 6.82 & 3.87 & \textbf{3.68} & 6.34 & {16.02}  \\   
    \bottomrule
    \end{tabular}
\end{footnotesize}     
\end{minipage}
\caption{\textbf{Depth edges help prioritize learning} of texture discontinuities over geometric ones. We report the RMS deviation from a plane (\underline{lower is better}) for the reconstructed checkerboard surfaces in mm. We note that \adashell performs slightly worse than volumetric methods \volsdf and \neus.  Except for \unisurf, all improve the quality of the surface measured with only stereo. Qualitative results are shown in \cref{fig:unisurf_pathology}.}     
\label{tab:jointly_refining_shape_appearance}
\end{table}
\begin{figure*}
\centering
\includegraphics[width=0.8\textwidth]{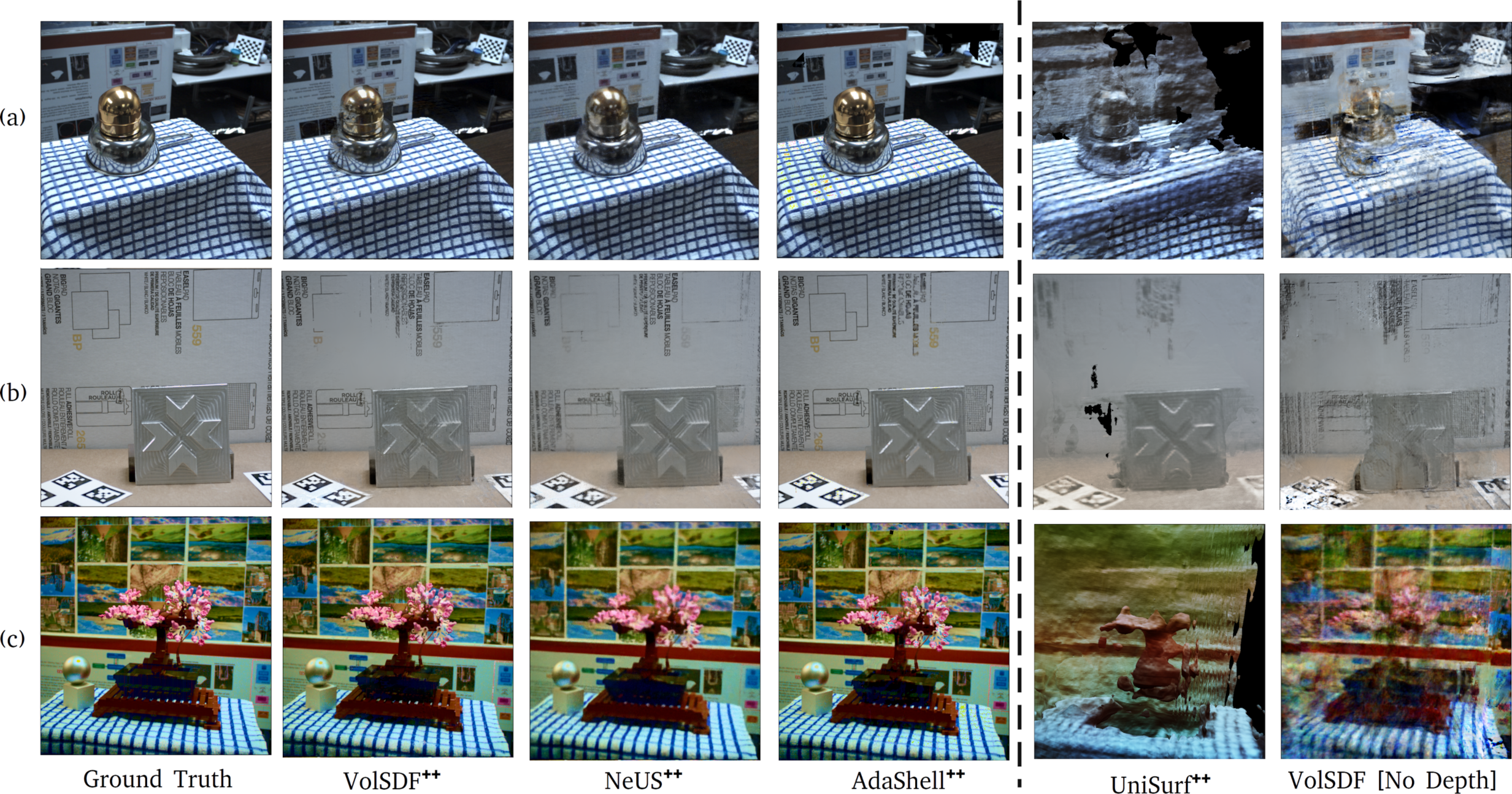}
\caption{\textbf{Relative performance of the baselines}. Quantitative results in \cref{tab:training_speed}, discussions in \cref{sc:baselines}.}
\label{fig:methods_comparison}
\end{figure*}
\subsection{View synthesis with dense depth}\label{sc:experiment_performnce}
Incorporating dense metric depth and our sampling strategy from \cref{sc:sdf_with_depth,sc:appearance_and_shape} enables \adashell, \volsdf, and \neus to perform competitively across challenging scenes. Scene A (\cref{fig:methods_comparison}(a)) looks at a couple of reflective objects with large variation in view dependent appearance. Additionally, there are large local errors in the captured depth maps due to specularities in the scene. We capture six stereo pairs, train on 11 images and test on one image. Scene B (\cref{fig:methods_comparison}(b)) features a rough metallic object of relatively simple geometry  captured by a 16mm lens (450 mm focal length, shallow depth of field). We capture four stereo pairs, train on seven images and test on one image. Scene C (\cref{fig:methods_comparison}(c))  features a fairly complicated geometry and is captured with 12 stereo pairs. We train on 22 images and test on two. Quantitative results of our experiments are in \cref{tab:training_speed}. We observe that \adashell, which is roughly 15\% faster per gradient step than \volsdf, generally converges the fastest (wall clock time) to a target PSNR. When the geometry is very complicated (scene C), an equally complicated sampling volume negates the efficiency gains of our sampler. We could not find good parameters for \unisurf for any of these sequences. 
\newline \indent View synthesis was unsuccessful without the inclusion of dense depth. We trained \volsdf with no depth supervision (equivalent to \cite{yariv2021volsdf}) until saturation (less than 0.1 PSNR increase for 1000 consecutive epochs). The reconstructions, none of which had a PSNR of 18 or higher, are shown in the last column of \cref{fig:methods_comparison}.
\begin{table}   
\begin{minipage}{\columnwidth}
    \centering
    \begin{footnotesize}
    \begin{tabular}{@{}l||c|c|c@{}}
    \toprule
          Metric &  \volsdf  & \neus & \adashell  \\
          \hline
          27.5+  & 21.3 \; 33.1 \; 40.0 & 70.5 \; 100+ \; 100+ & 22.6  \; 23.2  \; 94.6  \\ 
    \hline
          100K & 30.28\; 30.33 \; 30.69 & 27.82\;29.45\;25.31 & 31.56 ;\ 31.45 \; 28.27  \\
          \bottomrule
    \end{tabular}
\end{footnotesize}      
\end{minipage}
\caption{\textbf{Training performance for view synthesis.} We report two metrics - number of steps required to reach or exceed a PSNR of 27.5 and PSNR at the end of 100K gradient steps. We observe that all the baselines perform competitively and ignoring depth and additional supervision signals (last two columns) leads to failures in the view synthesis tasks.} \label{tab:training_speed}
\end{table}
\subsection{Using noisy depth}\label{sc:experiment_noisy_depth}
\begin{table}[b]
\begin{minipage}{\columnwidth}
    \centering
    \begin{footnotesize}
    \begin{tabular}{@{}l||ccc@{}}
    \toprule
          scene & \cref{fig:schematic_of_adashell}(a)[5] &  \cref{fig:methods_comparison}(b)[7] & \cref{fig:unisurf_checker_vertical}[5]   \\
          \hline
          edge sampling & \textbf{491} & \textbf{403} & \textbf{225}   \\ 
          no edge sampling  & 593 & 419 & 251    \\  
          noisy stereo  & 600 & 523 & 369    \\  
        \midrule
         27.5+  \adashell w/ noise & 7.93 & 20.2 & 5.12     \\   
         27.5+  \adashell w/o noise & 7.85 & 23.2 & 2.71    \\   
         \textbf{25.0+} \neus w/ noise & {49.4} & {36.0} & {32.9}    \\   
    \bottomrule
    \end{tabular}
\end{footnotesize}      
\end{minipage}
\caption{\textbf{Effect of noisy depth and depth edges.} Top: The surface reconstruction quality (Hausdorff distance, \underline{lower is better}) with conventional (noisy) stereo compared with surface recovered by \neus on learned stereo. Bottom: gradient steps (in 1000s \underline{lower is faster}) required to surpass a test time target PSNR. We specify the count of training views in $[~]$ braces.}     
\label{tab:effect_of_noise_and_depth_edge}
\end{table}
To investigate the effects of noise in the depth maps, we obtain the depths of scenes using conventional stereo. We used semi-global matching stereo (\cite{hirschmuller2005accurate}) with a dense census cost (\cite{zabih1994non}) and sub-pixel refinement on tone mapped HDR images to calculate the surface depth.  Surface normals were calculated using the spatial gradients of the depth maps. 
To focus on the performance of our approaches, we did not filter or smooth the depth obtained from conventional stereo. From the top of \cref{tab:effect_of_noise_and_depth_edge}, we observe that \neus strictly improves the quality of the surface reconstructed from just noisy stereo (row 1 and 2 versus row 3), especially when edge sampling is enabled. 
If the end goal is just view synthesis, \adashell, which blends the advantages of volumetric and surface based rendering, performs equally well with large noise in depth, whereas \neus takes many more iterations to converge. This indicates that photorealistic view synthesis with a volumetric renderer is possible with noisy depth data. However conventional stereo often introduces large local errors which our approaches were unable to improve significantly. 
\newline \indent In the presence of noisy depth, the quality of the reconstructed surface was enhanced through edge-based sampling (\cref{sc:appearance_and_shape,eq:edge_selection_probabililty}). Our sampling strategy allocated samples away from depth edges, where the noise was more prevalent, leading to fewer gradient steps spent modelling areas with higher noise. \Cref{tab:effect_of_noise_and_depth_edge} presents the quantitative details of the experiment.
\begin{figure*}
\centering
\includegraphics[width=0.85\textwidth]{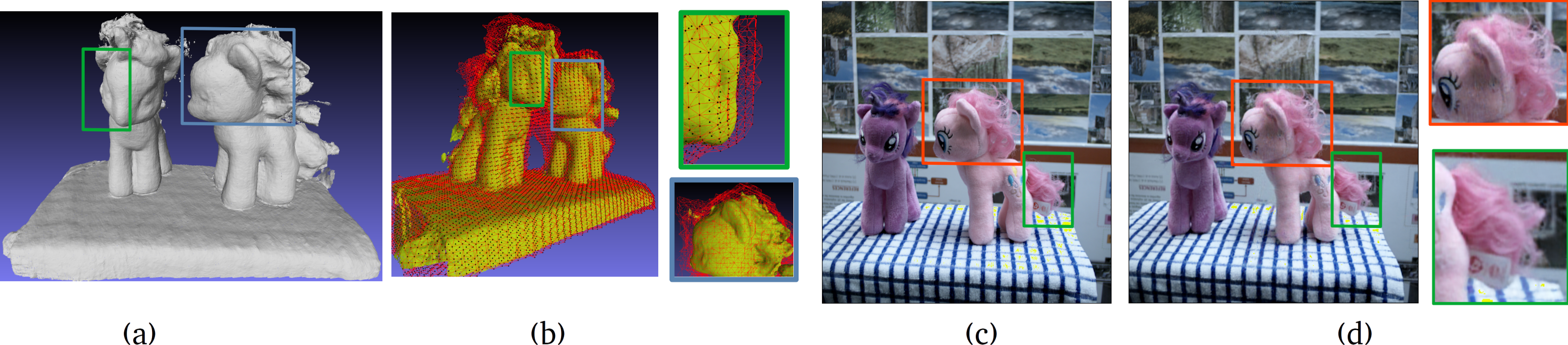}
\caption{\textbf{\adashell recovers sampling volumes} similar to \cite{adaptiveshells2023}. Fig. a shows the geometry recovered with 5 RGBD tuples. Figures b displays the sampling volumes around the geometry after \adashell has converged -- we note the similarity of this step with \cite{adaptiveshells2023}. Figs. c and d are the ground-truth and reconstructed test images. \adashell combines the advantages of volumetric rendering (see insets in fig. d) and surface based rendering (fig. b). More details are in the supplementary materials.}
\label{fig:schematic_of_adashell}
\end{figure*}
\subsection{Relighting}\label{sc:experiment_BRDF_capture}
\begin{figure*}[htbp]
    \begin{subfigure}[b]{0.15\textwidth}
    \centering
    \includegraphics[width=0.95\textwidth]{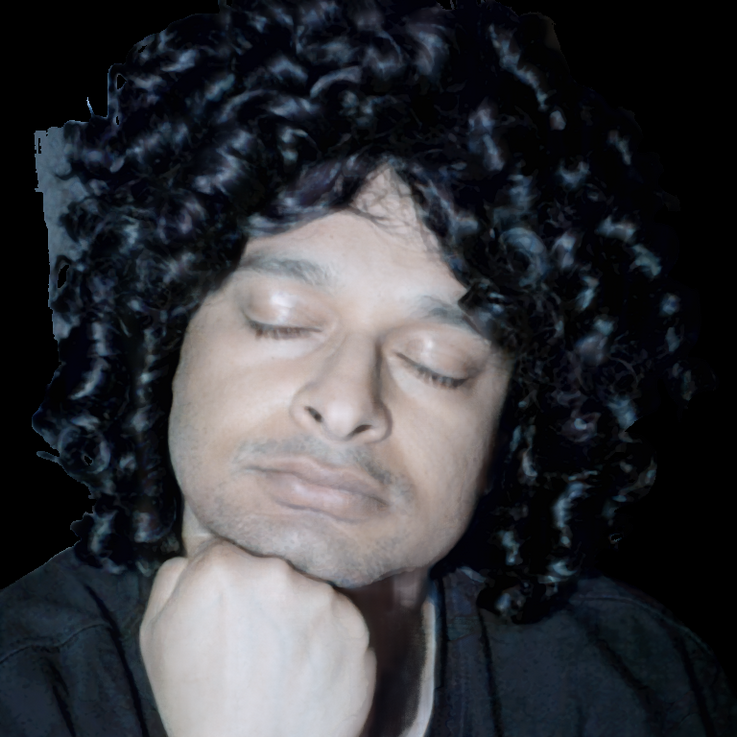}
    \caption{Face 1 (\texttt{F\_1})}
    \label{fig:face_1_relit}
    \end{subfigure}
        \begin{subfigure}[b]{0.15\textwidth}
    \centering
    \includegraphics[width=0.95\textwidth]{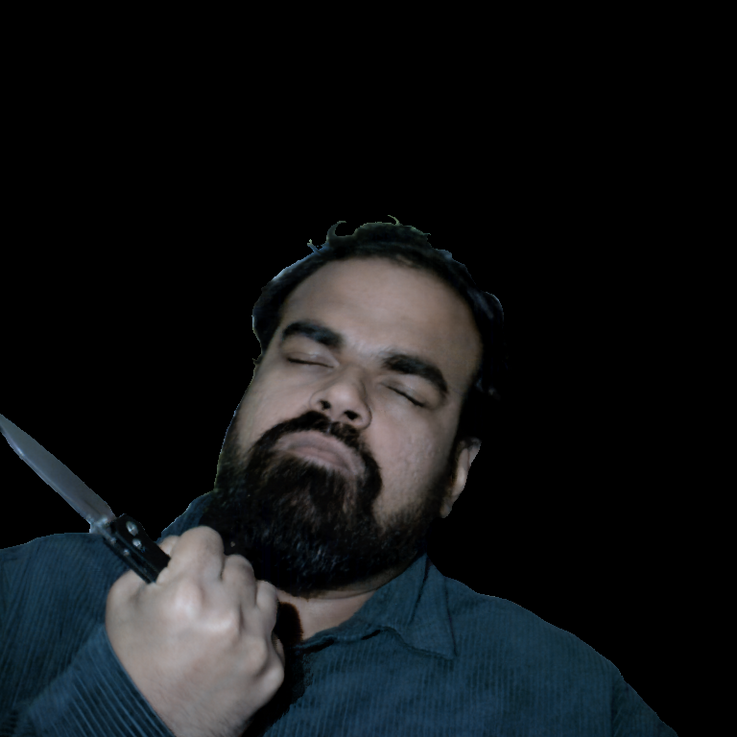}
    \caption{ Face 2 (\texttt{F\_2})}
    \label{fig:face_2_relit}
    \end{subfigure}
        \begin{subfigure}[b]{0.15\textwidth}
    \centering
    \includegraphics[width=0.95\textwidth]{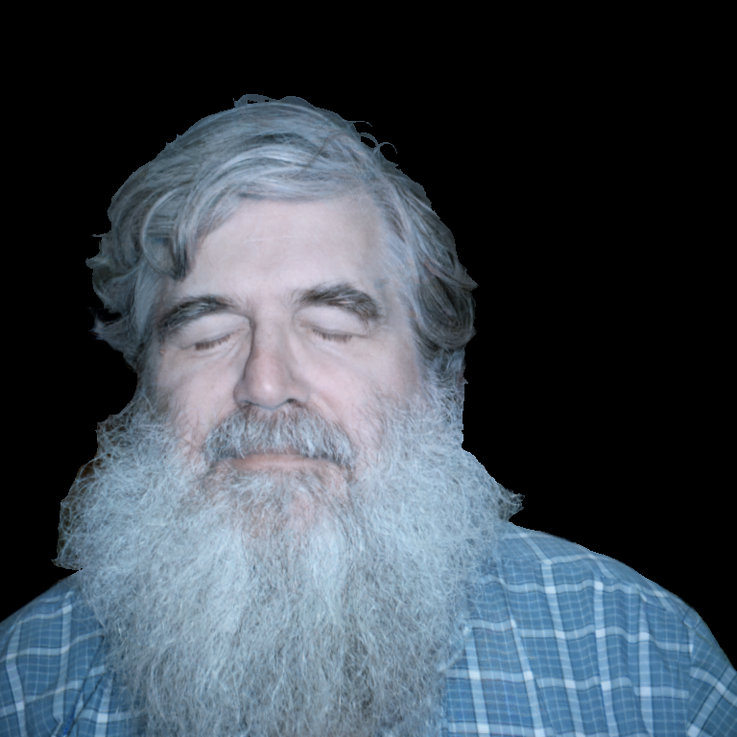}
    \caption{Face 3 (\texttt{F\_3})}
    \label{fig:face_3_relit}
    \end{subfigure}
    \begin{subfigure}[b]{0.15\textwidth}
    \centering
    \includegraphics[width=0.95\textwidth]{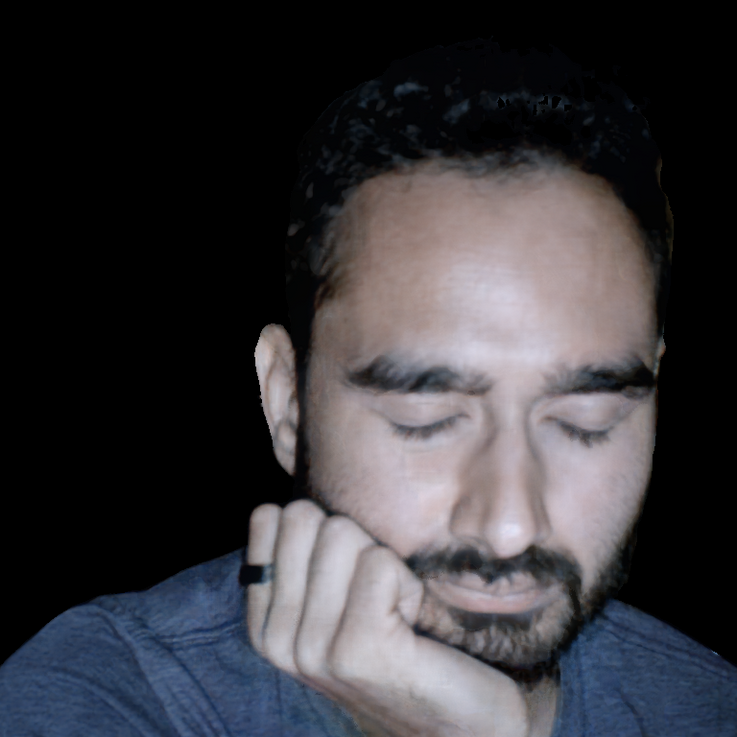}
    \caption{Face 4(\texttt{F\_4})}
    \label{fig:face_4_relit}
    \end{subfigure}
        \begin{subfigure}[b]{0.15\textwidth}
    \centering
    \includegraphics[width=0.95\textwidth]{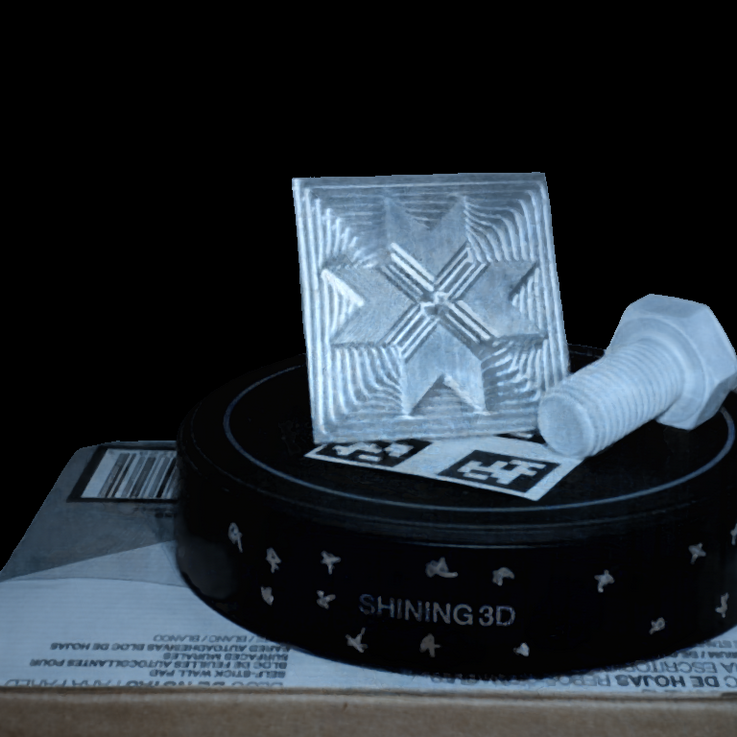}
    \caption{Shiny 1 (\texttt{S\_1})}
    \label{fig:shiny_1_relit}
    \end{subfigure}
    \begin{subfigure}[b]{0.15\textwidth}
    \centering
    \includegraphics[width=0.95\textwidth]{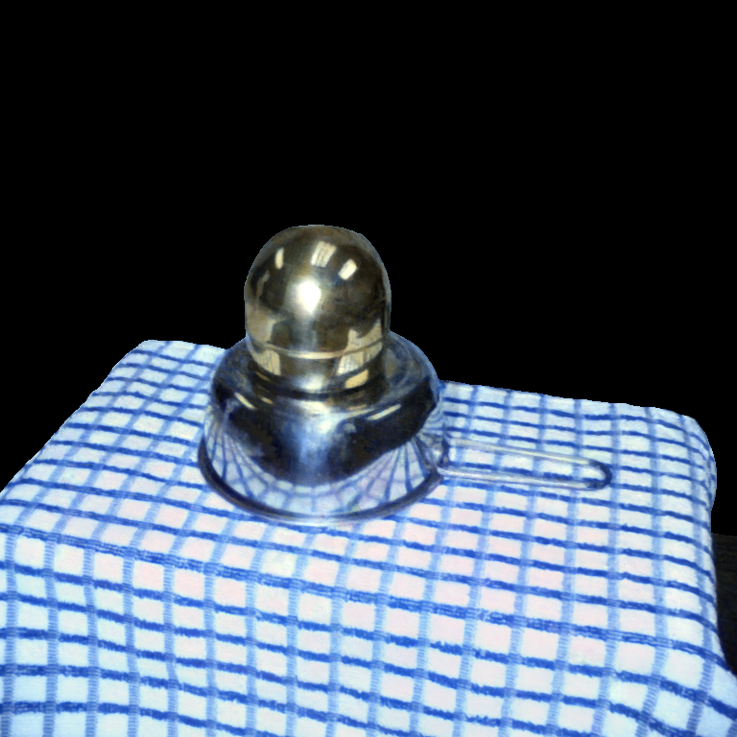}
    \caption{ Shiny 2 (\texttt{S\_2})}
    \label{fig:shiny_2_relit}
    \end{subfigure}
\caption{\textbf{Relighting scenes} can be achieved with \adashell trained with multi-illumination images. Discussion and results in \cref{sc:experiment_BRDF_capture,tab:relighting_expts}. We capture 12 flash lit images for all the camera views and we use alternate flashes for all the training views (6 per view). Figures above show one (of six) flash configurations for the test view. More results on the project website. }
\label{fig:relighting_main}
\end{figure*}
\begin{table}   
\begin{minipage}{\columnwidth}
    \centering
    \begin{footnotesize}
    \begin{tabular}{@{}l||cccccc@{}}
    \toprule
          Scene & \texttt{F\_1} &  \texttt{F\_2} & \texttt{F\_3} & \texttt{F\_4} &\texttt{S\_1}  & \texttt{S\_2}   \\
          \hline
          mask & \textbf{28.95} & \textbf{31.19} & \textbf{29.56} & \textbf{27.28} & \textbf{25.18} & \textbf{24.51}  \\ 
          no mask  & 27.17 & 30.05 & 27.82 & 25.46 & 23.65 & 21.88   \\   
    \bottomrule
    \end{tabular}
\end{footnotesize}     
\end{minipage}
\caption{\textbf{Relighting scenes} with a volumetric renderer. We report PSNR (\underline{higher is better}) under two heads -- masked and unmasked relit images, to offset the effects of incorrect shadows cast on the background. The unmasked reconstructions generally have a poorer PSNR because our implicit scene understanding approach does not approximate a ray tracer and cannot cast correct shadows on the background. The lower PSNR for reconstructing the shiny objects is mainly due to the inability of the network to model saturation caused by reflection. Results in \cref{fig:relighting_main}. }     
\label{tab:relighting_expts}
\end{table}
We capture multi-illumination images with known light poses and recover geometry independently of appearance. This allows us to infer the illumination dependent appearance using a combination of physically based appearance parameters -- e.g. the Disney Principled BRDF\cite{burley2012physically}. As a benchmark, we upgraded the closest related work, \cite{cheng2023wildlight}, which uses the full gamut of the Disney BRDF parameters, with \neus, to incorporate dense depth. For the data we collected, the optimization process  as implemented by \cite{cheng2023wildlight}, was quite brittle and some parameters (e.g. `clearcoat-gloss') would often take precedence over other appearance parameters (e.g. `specular-tint') and drive the optimization to a poor local minima. We demonstrate this problem in detail in the supplementary material. We found the optimization of a subset of appearance parameters (`base-color', `specular-tint', and `roughness') to be the most stable. \cite{zhang2022iron,brahimi2024supervol} conclude the same. 
\newline \indent For relighting, we explore two avenues -- the inference step of our approach as a volumetric renderer and a mesh created with the appearance parameters as texture (\cref{fig:teaser}). Quantitative and qualitative results of the volumetric renderer are shown in \cref{tab:relighting_expts,fig:relighting_main}. We used \cite{blender} to unwrap the geometry and generate texture coordinates whose quality exceeded \cite{xatlas} and our implementation of \cite{srinivasan2023nuvo}. None of our approaches worked on the ReNe dataset (\cref{tab:dataset_comparison}, \cite{Toschi_2023_RENE}) due to low view diversity, and the absence of metric depth. We used the labels in \cref{fig:mfc_specularity_labels} to allocate more gradient steps for learning the regions with higher appearance variation. We provide more details in the supplementary material.
\section{Limitations} 
Although we achieve state of the art results in view-synthesis and relighting with a few views, our approach struggles to represent transparent objects and accurately capture the geometry of reflective surfaces. \cite{liu2023nero} address the problem of reflective objects by modelling background reflections and is based on the architecture proposed by \cite{wang2021neus}. As \neus enables \cite{wang2021neus} to use possibly noisy metric depth, it can potentially be extended to model reflective objects. 
\newline \indent Our approaches require metric depth and depth edges for the best performance. Our approach relies on capture devices with reasonable quality depth measurements. Future work will address incorporation of monocular and sparse depth priors with depth edges. 
\newline \indent Incorporation of metric depth introduces a strong bias, often limiting super resolution of geometry sometimes achieved in neural 3D scene representation (see e.g. \cite{li2023neuralangelo}). Decreasing the effect of \cref{eq:igr_loss} during training may potentially encourage geometric superresolution and is future work. 
\newline \indent Finally, modern grid based representations (see e.g. \cite{reiser2023merf,duckworth2023smerf}) produce very compelling view interpolation results at a fraction of the computational cost of a state of the art volumetric renderer (e.g. \cite{adaptiveshells2023,muller2022instantngp}). However, they need to be ``distilled'' from a pre-trained volumetric view interpolator. Future work can investigate the use of depth priors to train a grid based representation directly from color and depth images.  
\section{Conclusions}
We present a solution to incorporate dense metric depth into neural 3D reconstruction which enables state of  the art geometry reconstruction. We examine a corner case of jointly learning appearance and geometry and address it by incorporating additional supervision signals. Additionally, we describe a variant of the multi-flash camera to capture the salient supervision signals needed to improve photorealistic 3D reconstruction and demonstrate a pipeline for view synthesis and relighting of small scenes with a handful of training views.

{
    \small
    \bibliographystyle{ieeenat_fullname}
    \bibliography{main}
}
\clearpage
\maketitlesupplementary

This supplementary document inherits the figure, equation, table, and reference numbers from the main document. Additional results may be viewed at \href{https://stereomfc.github.io}{https://stereomfc.github.io}. 

\section{Representations and implementation details}\label{sc:implementation_details}
Our scene representation consists of two networks -- an intrinsic network $\mathcal{N}(\theta)$ and an appearance network $\mathcal{A}(\phi)$. We follow \cite{li2023neuralangelo} to build and train $\mathcal{N}(\theta)$. We use 18 levels of hashgrid encodings \cite{muller2022instantngp} to encode the input and a two layer (128 neurons/layer) MLP to generate the intrinsic embedding. The first channel of the embedding, $\mathcal{S}(\theta)$ is trained with \cref{eq:igr_loss} to recover a signed distance field of the scene as described in \cref{sc:sdf_with_depth}. The rest of the 127 channels of the embedding $\mathcal{E}(\theta)$ are passed on to the appearance network $\mathcal{A}(\phi)$ as an input. 
\newline \indent The appearance network takes $\mathcal{E}(\theta)$, the viewing direction (encoded with 6 levels of sinusoidal encodings following \cite{tancik2020fourier}), and optionally the illumination direction (if recovering BRDF) to generate colors. The neural network is built with 2 layers of fully connected MLPs (128 neurons/layer) with skip connections. 
\newline \indent The neural signed distance field $\mathcal{S}(\theta)$ is optimized to return the signed distance of a point  from its nearest surface $\mathcal{S}(\theta): \mathbb{R}^3 \to \mathbb{R}$. 
The surface of the object can be obtained from the zero-level set of $\mathcal{S}(\theta)$ -- i.e. for all surface points $\mathbf{x}_s \in \mathbb{R}^3 ~ | ~\mathcal{S}(\mathbf{x}_s|\theta) = 0$. We train $\mathcal{S}(\theta)$ by minimizing a geometric loss $\ell_D$ (\cref{eq:igr_loss}). We follow \cite{yariv2021volsdf} to transform the  distance of a point $\vec{p}_i = \vec{r}|_{t_i}$ in a ray to its closest surface $s_i = \mathcal{S}(\theta,\vec{p}_i)$ to the scene density (or transmissivity).
\begin{equation}\label{eq:volsdf_transform}
    \Psi_\beta(s) = \begin{cases}
                    0.5\exp(\frac{s}{\beta}), & s\leq0\\
                    1-0.5\exp(\frac{-s}{\beta}), & \text{otherwise}.
                    \end{cases}
\end{equation}
\newline \indent
To render the color $\mathbf{C}$ of a single pixel of the scene at a target view with a camera centered at $\vec{o}$ and an outgoing ray direction $\vec{d}$, we calculate the ray corresponding to the pixel $\vec{r} = \vec{o} + t\vec{d}$, and sample a set of points $t_i$ along the ray. The networks $\mathcal{N}(\theta)$ and $\mathcal{A}(\phi)$ are then evaluated at all the $\mathbf{x}_i$ corresponding to $t_i$ and the per point color $\mathbf{c}_i$. The transmissivity $\tau_i$ is obtained and composited together using the quadrature approximation from \cite{max1995optical} as:
\begin{equation}\label{eq:nerf_volumetric_rendering}
    \mathbf{C} = \sum_{i}\rm{exp}( - \sum_{j<i} \tau_j \delta_j ) (1-\rm{exp}(-\tau_j\delta_j))\mathbf{c}_i,   \; \; \; \delta_i = t_i - t_{i-1}
\end{equation}
The appearance can then be learned using a loss on the estimated and ground truth color $\mathbf{C}_{gt}$
\begin{equation}\label{eq:color_loss}
    \ell_{C} = \mathbb{E}\left[||\mathbf{C} - \mathbf{C}_{gt}||^2\right]
\end{equation}
The appearance and geometry are jointly estimated by minimizing the losses in \cref{eq:total_reconstruction_loss} using stochastic gradient descent \cite{kingma2014adam}. 
\begin{equation}\label{eq:total_reconstruction_loss}
    \ell = \ell_C + \lambda_{g} \ell_D + \lambda_{c} \mathbb{E}(|\nabla^2_\mathbf{x} \mathcal{S}(\mathbf{x}_s)|)
\end{equation}
$\lambda$s are hyperparameters and the third term in \cref{eq:total_reconstruction_loss} is the mean surface curvature minimized against the captured surface normals. As the gradients of the loss functions $\ell_C$ and $\ell_D$ propagate through $\mathcal{A}$ and $\mathcal{N}$ (and $\mathcal{S}$ as it is part of  $\mathcal{N}$) the appearance and geometry are learned together. 
\subsection{Details of our baselines}
\begin{figure*}
\centering
\begin{subfigure}[b]{0.145\textwidth}
\centering
\includegraphics[width=0.90\textwidth]{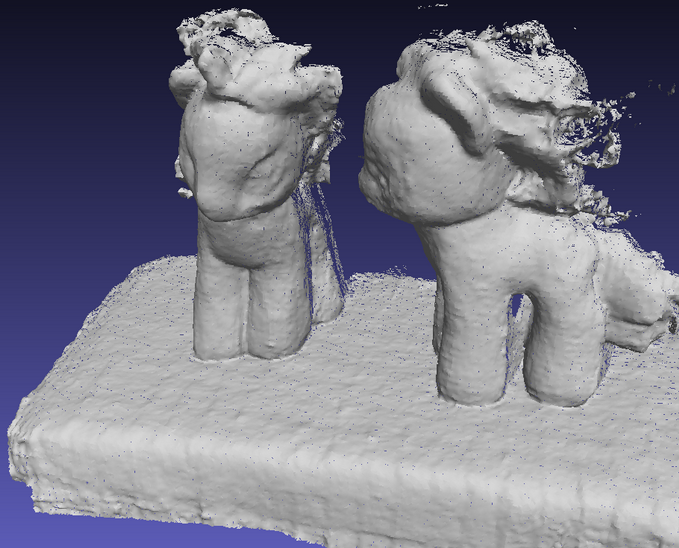}
\caption{}
\label{fig:ponies_clean_mesh}
\end{subfigure}
\begin{subfigure}[b]{0.145\textwidth}
\centering
\includegraphics[width=0.92\textwidth]{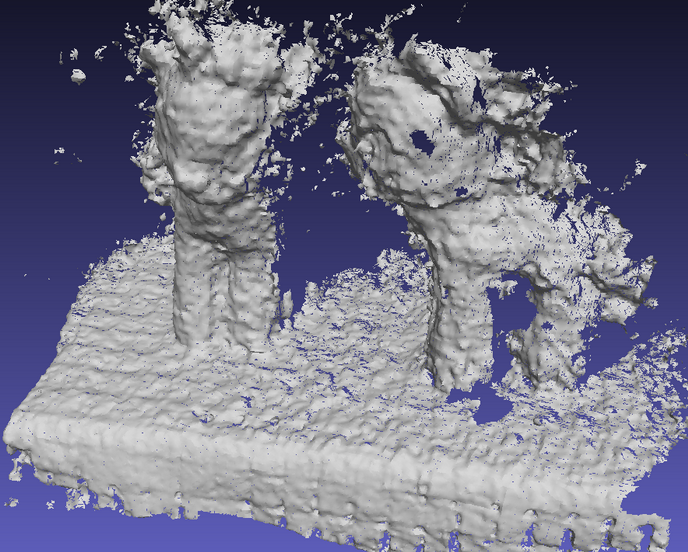}
\caption{}
\label{fig:ponies_noisy_mesh}
\end{subfigure}
\begin{subfigure}[b]{0.145\textwidth}
\centering
\includegraphics[width=0.90\textwidth]{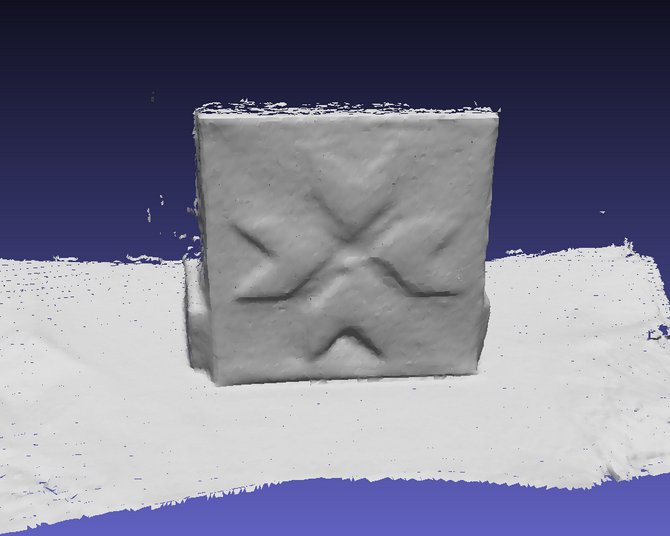}
\caption{}
\label{fig:al_clean_mesh}
\end{subfigure} 
\begin{subfigure}[b]{0.145\textwidth}
\centering
\includegraphics[width=0.90\textwidth]{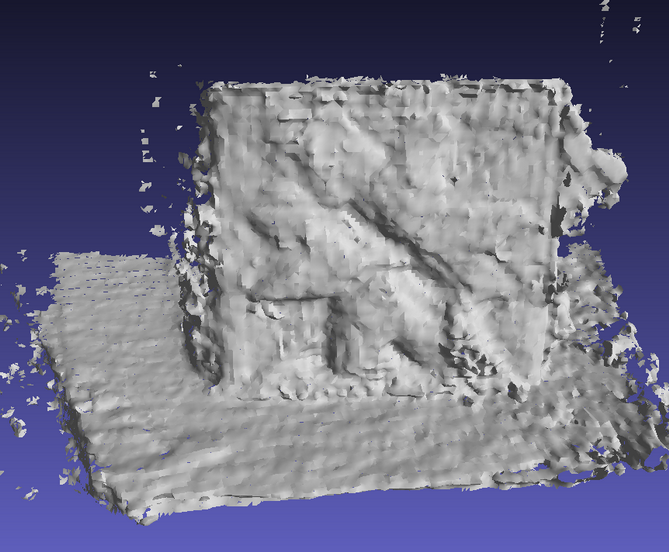} 
\caption{}
\label{fig:al_noisy_mesh}
\end{subfigure}
\begin{subfigure}[b]{0.145\textwidth}
\centering
\includegraphics[width=0.90\textwidth]{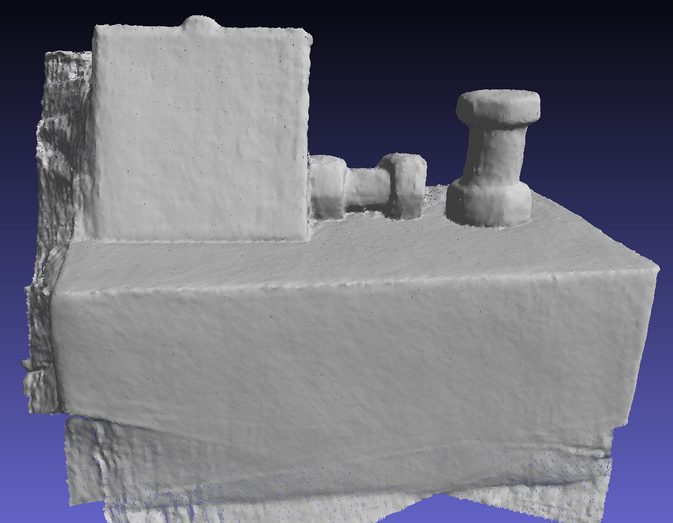} 
\caption{}
\label{fig:checker_clean_mesh}
\end{subfigure}
\begin{subfigure}[b]{0.145\textwidth}
\centering
\includegraphics[width=0.90\textwidth]{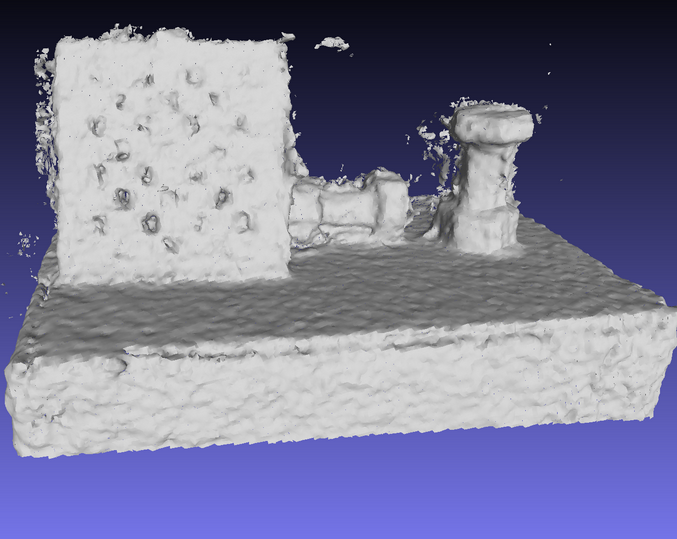} 
\caption{}
\label{fig:checker_noisy_mesh}
\end{subfigure}
\caption{\textbf{\adashell can work with noisy depth} with little loss in view interpolation performance and training time. However, the surface recovered is also noisy. \Cref{fig:ponies_clean_mesh,fig:al_clean_mesh,fig:checker_clean_mesh} are the geometries recovered with no noise in depth and \cref{fig:ponies_noisy_mesh,fig:al_noisy_mesh,fig:checker_noisy_mesh} are with noisy depths. Although \adashell does not de-noise the geometry, there is very little degradation in performance (training speed and view synthesis quality) with noisy depth. \neus struggles at the task of view synthesis with noisy depth. Details in  \cref{sc:experiment_noisy_depth,tab:effect_of_noise_and_depth_edge}. }
\label{fig:using_noisy_normals}
\end{figure*}
We implemented four baselines to investigate the effects of incorporating dense metric depth and depth edges into neural view synthesis pipelines. 
\newline \indent \textbf{\volsdf} is our method similar to VolSDF~\cite{yariv2021volsdf} and MonoSDF\cite{Yu2022MonoSDF}. We represent the scene with $\mathcal{N}$ and $\mathcal{A}$ and train it with metric depth and color by minimizing \cref{eq:total_reconstruction_loss}. The samples for \cref{eq:color_loss} are drawn using the ``error-bounded sampler'' introduced by \cite{yariv2021volsdf}. 
\newline \indent \textbf{\neus} represents a modified version of NeUS~\cite{wang2021neus}, where we use the training schedule and structure of $\mathcal{N}$ from \cite{li2023neuralangelo}, the appearance network $\mathcal{A}$ is adopted from NeUS and we optimize \cref{eq:igr_loss} along with \cref{eq:color_loss}. In addition to $\mathcal{A}$, \neus also has a small 4 layer MLP (32 neurons per layer) to learn the radiance of the background as recommended in the original work by \cite{wang2021neus}. 
\newline \indent \textbf{\unisurf} is our method inspired by UniSurf\cite{oechsle2021unisurf}. We represent the scene's geometry using a pre-optimized implicit network $\mathcal{N}$ as outlined in \cref{sc:sdf_with_depth}. We follow the recommendations of \cite{oechsle2021unisurf} to optimize $\mathcal{A}$. UniSurf exposes a hyperparameter to bias sampling of \cref{eq:nerf_volumetric_rendering} towards the current estimate of the surface. As we pre-optimize the surface, we can find the surface point $\mathbf{x}_s = \mathbf{o}+t_s\mathbf{d}$ through sphere tracing $\mathcal{S}$ along a ray. The intersection point $t_s$ can then be used to generate $N$ samples along the ray to optimize \cref{eq:color_loss}.  
\begin{align}\label{eq:unisurf_sampler}
t_i = \mathcal{U}\left[ t_s + \left( \dfrac{2i-2}{N} - 1\right)\Delta,  t_s + \left( \dfrac{2i}{N} - 1\right)\Delta\right]
\end{align}
\Cref{eq:unisurf_sampler} is the distribution used to draw samples and $\Delta$ is the hyperparameter that biases the samples to be close to the current surface estimate. We optimize $\mathcal{S}$ independent of \cref{eq:nerf_volumetric_rendering} by just minimizing \cref{eq:igr_loss} with registered depth maps (see \cref{sc:sdf_with_depth}). We use this method to study the effects of volumetric rendering versus surface rendering. We found this strategy to be very sensitive to the hyperparameter $\Delta$ and its decay schedule as the training progressed. While best parameters for some sequences resulted in very quick convergence, they were very hard to come across and generally, poorer choices led to undesirable artifacts (see e.g. \cref{fig:unisurf_pathology}). 
\newline \indent We describe \textbf{\adashell} in the following section. 
\subsection{\adashell: Accelerating training with dense depth}\label{sc:adashell}
\begin{figure*}
\centering
\includegraphics[width=0.85\textwidth]{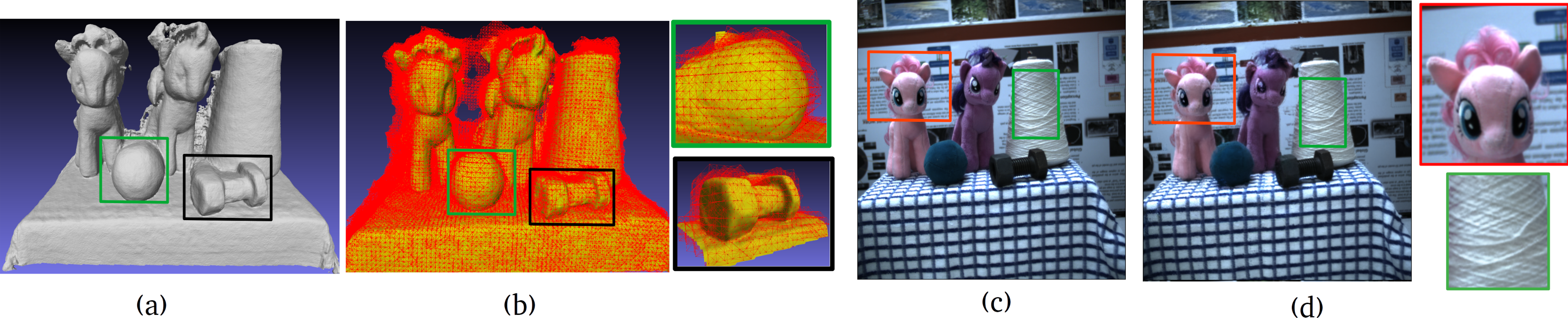}
\caption{\textbf{Another example of shells recovered by \adashell}. Fig. a shows the geometry recovered with 5 RGBD tuples. Figure b displays the sampling volumes around the geometry after \adashell has converged -- we note the similarity of this step with \cite{adaptiveshells2023}. Figs. c and d are the ground-truth and reconstructed test images. Related example in \cref{fig:schematic_of_adashell}.}
\label{fig:schematic_of_adashell_2}
\end{figure*}
The slowest step in training and inference for neural volumetric representations is generally the evaluation of \cref{eq:nerf_volumetric_rendering}. In this section we describe our method to accelerate training by incorporating metric depth. \newline \indent A method to make training more efficient involves drawing the smallest number of the most important samples of $t_i$ for any ray. The sampling of $t_i$ is based on the current estimate of the scene density and although these samples  can have a large variance, given a large number of orthogonal view pairs (viewpoint diversity), and the absence of very strong view dependent effects, the training procedure is expected to recover an unbiased estimate of the true scene depth (see e.g. \cite{BayesRaysGoli2023}). We can accelerate the convergence by a) providing high quality biased estimate of the scene depth and b) decreasing the number of samples for $t_i$ along the rays. 
\newline \indent Given the high quality of modern deep stereo (we use\cite{xu2022gmflow}) and a well calibrated camera system, stereo depth can serve as a good initial estimate of the true surface depth. We use stereo depth, aligned across multiple views of the scene to pre-optimize the geometry network $\mathcal{S}(\theta)$. The other channels $\mathcal{E}(\theta)$ of $\mathcal{N}$ remain un-optimized. A pre-optimized $\mathcal{S}$ can then be used for high quality estimates of ray termination depths. 
\newline \indent  \cite{yariv2021volsdf,oechsle2021unisurf,wang2021neus} recommend using root finding techniques (e.g. bisection method) on scene transmissivity (\cref{eq:volsdf_transform}) to estimate the ray termination depth. The samples for \cref{eq:nerf_volumetric_rendering} are then generated around the estimated surface point. Drawing high variance samples as $\mathcal{N}$ and $\mathcal{A}$ are jointly optimized reduces the effect of low quality local minima, especially in the initial stages of the optimization. As we have a pre-trained scene transmissivity field ($\mathcal{S}$ transformed with \cref{eq:volsdf_transform}), we can draw a few high-quality samples to minimize the training effort. 
\newline \indent We found uniformly sampling  around the estimated ray-termination depth (\unisurf baseline in \cref{sc:baselines,fig:unisurf_pathology}) to be unsuitable. Instead, we pre-calculated a discrete sampling volume by immersing $\mathcal{S}$ in an isotropic voxel grid and culling the voxels which report a lower than threshold scene density. We then used an unbiased sampler from \cite{yariv2021volsdf} to generate the samples in this  volume. This let us greatly reduce the number of root-finding iterations and samples, while limiting the variance by the dimensions of the volume along a ray. As the training progresses, we decrease the culling threshold to converge to a thinner sampling volume around the surface while reducing the number of samples required.
\newline \indent  We show the sampling volume (at convergence) and our reconstruction results in \cref{fig:schematic_of_adashell,fig:schematic_of_adashell_2} respectively. We retain the advantages of volumetric scene representation as demonstrated by the reconstruction of the thin structures in the scene, while reducing training effort. We dub our method \adashell to acknowledge  \cite{adaptiveshells2023}, which demonstrates a related approach to accelerate inference. 
\newline \indent The ``shells'' shown in \cref{fig:schematic_of_adashell,fig:schematic_of_adashell_2} and the shells recovered by \cite{adaptiveshells2023} for small scenes are physically similar quantities. \cite{adaptiveshells2023} dilate and erode the original level-set of the scene (approximated by $\mathcal{S}$) using a hyperparameter. Our ``shells'' are also jointly estimated with the geometry as the training progresses. \cite{adaptiveshells2023} estimate the fall-off of the volume density values along a ray to determine the hyperparameters, which in turn determines the thickness of the ``shell''. They subsequently use uniform sampling (similar to \cref{eq:unisurf_sampler}, where the $\Delta$ now denotes the local thickness of the shell) to generate samples for rendering. Our work takes a discrete approach by immersing the zero-level set (in form of pre-optimized $\mathcal{S}$) in a dense isotropic voxel grid and culling the voxels which have a lower volume density, according to a preset hyperparameter that determines the thickness of the shell. Once the shell has been estimated, we use a unbiased density weighted sampler (instead of a uniform sampler) to generate samples along the ray inside the shell. We roughly follow \cite{sun2022neuralreconW} to generate samples along a segment of the ray guided by the voxels it intersects. The spatial density of samples is inversely proportional to their distance from the estimated surface. We implement this using the tools from NerfStudio\cite{li2022nerfacc,nerfstudio}. Our sampling strategy is more robust to errors in estimated geometry (as shown in \cref{sc:experiment_noisy_depth,fig:using_noisy_normals}) than other approaches --notably \neus and the original work (Adaptive Shells \cite{adaptiveshells2023}) which is based on NeUS(\cite{wang2021neus}).  
\subsection{Training Details}
\begin{figure*}
\centering
\includegraphics[width=\textwidth]{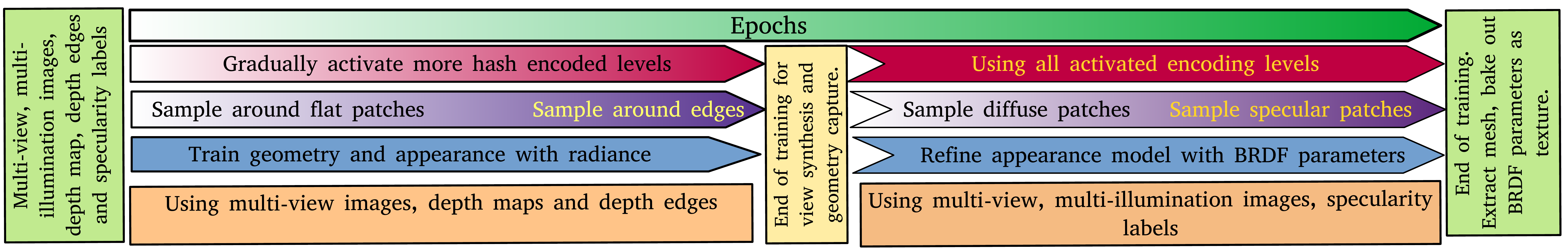}
\caption{\textbf{Our approach to recovering 3D assets from captured data}. In the first part, for \neus and \volsdf we jointly optimize geometry and appearance by minimizing \cref{eq:igr_loss,eq:color_loss}. For \adashell and \unisurf, we first optimize \cref{eq:igr_loss} for a fixed number of gradient steps before the joint optimization. At this stage the geometry is optimized and appearance is recovered as radiance. Following this, we use multi-illumination images with a truncated BRDF parametrization to refine the appearance model, given the geometry, to learn the reflectance parameters. }
\label{fig:schematic_of_end_to_end}
\end{figure*}
We ran our experiments on a Linux workstation with an Intel Core i9 processor, 64GB RAM, and an Nvidia RTX3090Ti graphics card with 25GB of vRAM. Across all the experiments for learning scene radiance, we implemented a hard cut-off of 100K gradient steps amounting to less than 4.5 hours of training time across all the experiments. 
\newline \indent Across all our baselines (\adashell, \volsdf, \neus, and \unisurf) we used the intrinsic network proposed in \cite{li2023neuralangelo}, with 2 layers of MLPs (128 neurons per fully connected layer) and 18 levels of input hash encodings activated gradually. Our input activation curriculum was based on the recommendations of \cite{li2023neuralangelo}, and was used jointly with our edge-aware sampling strategy \cref{sc:appearance_and_shape,eq:edge_selection_probabililty,fig:sampling_main} across all the scenes. 
The implementations of our baselines, design and bill-of-materials for the multi-flash camera system, dataset and the hyperparameters will be released soon. \Cref{fig:schematic_of_adashell} denotes the training steps graphically.
\subsection{Difference between our and prior work on neural scene understanding with depth}\label{sc:differences_in_methods}
\textbf{IGR\cite{gropp2020IGR}} was among the first to fit a neural surface to point samples of the surface. Our pipeline is largely inspired by that work. However, we have two main differences -- we use a smaller network, and periodically activate multi-resolution hash encodings as recommended by \cite{li2023neuralangelo} instead of using a fully connected set of layers with skip connections. Additionally, as we have access to depth maps, we identify the variance of the neighborhood of a point on the surface through a sliding window filter. We use this local estimate of variance in a normal distribution to draw samples for $\mathbf{x}_{s}^\Delta$ along each ray. Our strategy assumes that image-space pixel neighbors are also world space neighbors, which is incorrect along the depth edges. However, as the Eikonal equation should be generally valid in $\mathbb{R}^3$ for $\mathcal{S}$, the incorrect samples do not cause substantial errors and only contribute as minor inefficiencies in the pipeline. A more physically based alternative, following \cite{gropp2020IGR}, would be executing nearest neighbor queries at each surface point along the rays to estimate the variance for sampling. With about 80k rays per batch, $\sim$ 200K points in $(\mathbf{x}_s)$, and about 40k gradient steps executed till convergence, and a smaller network, our approach was more than two orders of magnitude faster than \cite{gropp2020IGR}, with no measurable decrease in accuracy of approximating the zero-level set of the surface. 
\newline \indent \textbf{NeuralRGBD\cite{Azinovic_2022_NeuralRGBD}} is the closest prior work based on data needed for the pipeline and its output. The scene is reconstructed using color and aligned dense metric depth maps. The authors aggregate the depth maps as signed distance fields and use the signed distance field to calculate weights for cumulative radiance along samples on a ray (\cref{eq:nerf_volumetric_rendering} in text). The weights are calculated with 
\begin{align}\label{eq:neural_rgbd_weights}
    w_i = \sigma\left(\dfrac{D_i}{tr}\right) \times \sigma\left(-\dfrac{D_i}{tr}\right)
\end{align}
where the $D_i$ is the distance to the surface point along a ray, and the truncation $tr$ denotes how fast the weights fall off away from the surface. \Cref{eq:neural_rgbd_weights} yields surface biased weights with a variance controlled by the parameter $tr$. Notably, the depth map aggregation does not yield a learned sign distance field (no Eikonal regularizer in the loss). The authors also include a `free-space' preserving loss to remove ``floaters''. As implemented, the pipeline needs the truncation factor to be selected per-scene. As the depth maps are implicitly averaged by a neural network, it is implicitly smoothed and therefore the pipeline is robust to local noise in the depth map.
\newline \indent \textbf{MonoSDF\cite{Yu2022MonoSDF}} is mathematically the closest prior method to our work and it uses dense scene depths and normals obtained by a monocular depth and normal prediction network (OmniData\cite{eftekhar2021omnidata}). MonoSDF defines the ray length weighted with the scene density as the scene depth $\mathbf{d}_{pred}$ and minimizes 
\begin{align}\label{eq:monosdf_depth_loss}
    \ell_{D} = \sum_r||\mathbf{wd}_{mono}+\mathbf{q} - \mathbf{d}_{pred}||_2^2
\end{align}
where $\{\mathbf{w,q}\}$ are scale and shift parameters. Estimating an affine transformation on the monocular depth $\mathbf{d}_{mono}$ is important because in addition to gauge freedom ($\mathbf{w}$), monocular depths also have an affine degree of freedom ($\mathbf{q}$). The scale and shift can be solved using least squares to align $\mathbf{d}_{mono}$ and $\mathbf{d}_{pred}$. The scene normals are calculated as gradients of $\mathcal{S}$ weighted with scene density along a ray.  Through a scale and shift invariant loss, MonoSDF calculates one set of $(\mathbf{w,q})$ for all the rays in the batch corresponding to a single training RGBD tuple.  In the earlier stages of the training, this loss helps the scene geometry converge. The underlying assumption is that there is an unique tuple $\{\mathbf{w,q}\}$ per training image that aligns $\mathbf{d}_{mono}$ to the actual scene depth captured by the intrinsic network $\mathcal{N}$. 
\newline \indent Our experiments with MonoSDF indicate that the network probably memorizes the set of $(\mathbf{w,q})$ tuples per training image. Explicitly passing an unique scalar tied to the training image (e.g. image index as proposed in \cite{martinbrualla2020nerfw}) speeds up convergence significantly. Success of MonoSDF in recovering both shape and appearance strongly depends on the quality of the monocular depth and normal predictions. Our experiments on using MonoSDF on the WildLight dataset(\cite{cheng2023wildlight}) or the ReNe dataset (\cite{Toschi_2023_RENE}) failed because the pre-trained Omnidata models performed poorly on these datasets. Unfortunately, as implemented, MonoSDF also failed to reconstruct scene geometry when the angles between the training views were small -- ReNe dataset views are maximally 45$^\circ$ apart. However, it demonstrates superior performance on the DTU and the BlendedMVS sequences while training with as low as three pre-selected views. Finally, our scenes were captured with a small depth of field and most of the background was out of focus, so the scene background depth was significantly more noisy than the foreground depth. We sidestepped this problem by assigning a fixed 1m depth to all the pixels that were in the background. Although this depth mask simplifies our camera pose estimation problem (by segregating the foreground from the background), it assigns multiple infeasible depths to a single background point. As we aggregate the depth maps into the intrinsic network ($\mathcal{N}$) by minimizing \cref{eq:igr_loss}, the network learns the mean (with some local smoothing) of the multiple depths assigned to the single background point. However, the scale and shift invariant loss is not robust to this and with masked depth maps, we could not reliably optimize MonoSDF on our sequences. We suspect that this is because the scale and shift estimates for each instance of \cref{eq:monosdf_depth_loss} on the background points yielded very different results, de-stabilizing the optimization.  
\newline \indent \textbf{\cite{roessle2022depthpriorsnerf}} and \textbf{\cite{kangle2021dsnerf}}
use sparse scene depth in the form of SfM triangulated points. \cite{roessle2022depthpriorsnerf} use learnt spatial propagation \cite{cheng2019convspatialprop} to generate dense depth maps from the sparse depth obtained by projecting the world points triangulated by SfM. \cite{kangle2021dsnerf} assign the closest surface depth at a pixel obtained by projecting the triangulated points to the image plane. Neither of these pipelines recover a 3D representation of the scene and focus on view synthesis using few views. 
\newline \indent \textbf{\cite{Sandstrom2023ICCV}} introduce a novel 3D representation -- ``Neural point clouds'' which includes geometric and appearance feature descriptors (small MLPs) grounded to a point in 3D. The geometry is recovered as the anchors of the ``neural points''. The appearance is calculated using a volumetric renderer which composits the outputs of the appearance descriptors of the neural points with the transmissivity of the neural points along the ray. The transmissivity of a neural point is calculated as a function of distances of a pre-set number of neighboring neural points. 
\subsection{Capturing approximate BRDF and generating textured meshes}\label{sc:BRDF_details}
\begin{figure*}
    \begin{subfigure}[b]{0.45\textwidth}
    \centering
    \includegraphics[width=\textwidth]{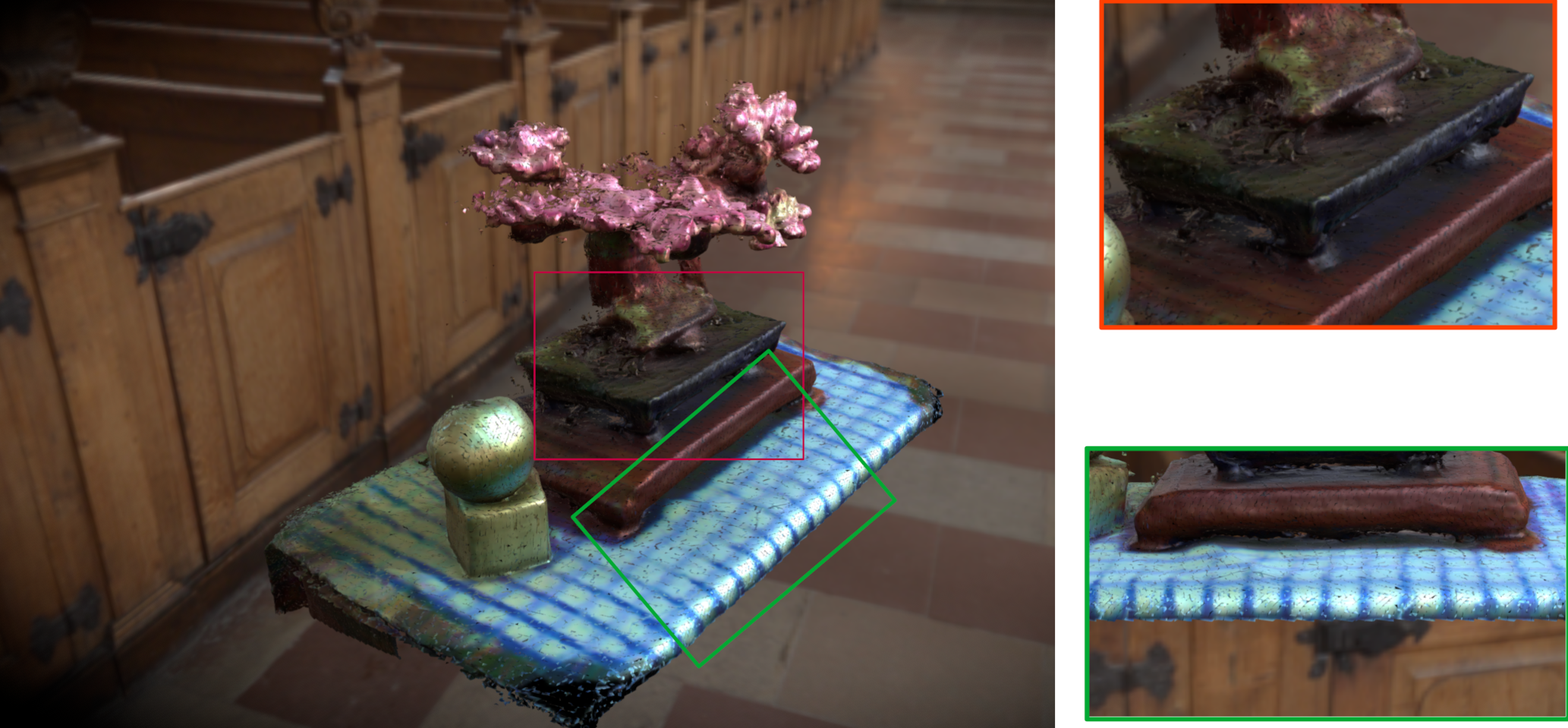}
    \caption{Limited parametrization}
    \label{fig:reduced_param_wildlight}
    \end{subfigure}
    \begin{subfigure}[b]{0.45\textwidth}
    \centering
    \includegraphics[width=\textwidth]{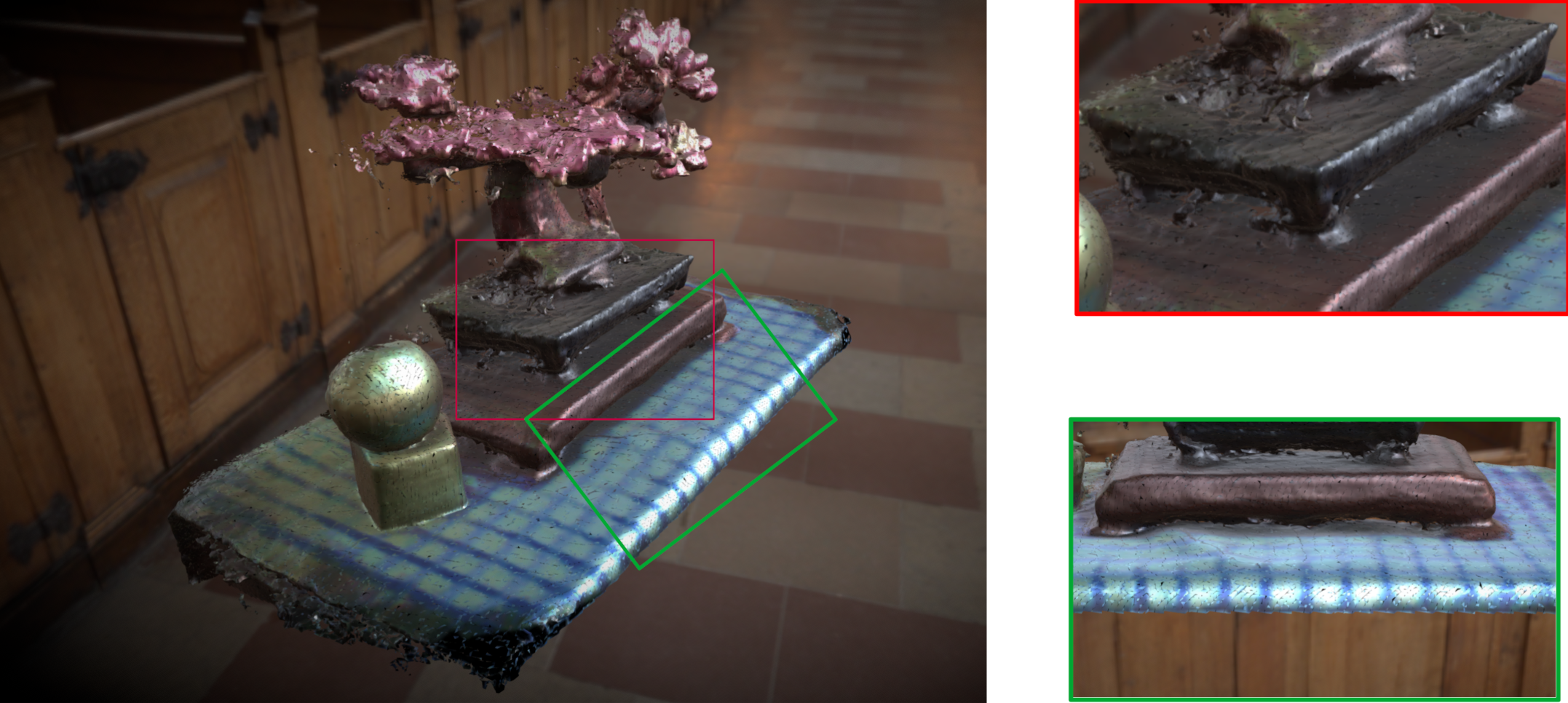}
    \caption{Full Disney BDRF parametrization}
    \label{fig:full_param_wildlight}
    \end{subfigure}
\caption{\textbf{Optimizing for the full Disney BRDF is difficult}. \Cref{fig:reduced_param_wildlight} shows our results with only specularity, roughness and metallic BRDF parameters. \Cref{fig:full_param_wildlight} depicts identical results utilizing the complete range of Disney BRDF parameters as outlined in \cite{cheng2023wildlight}. Note the excessive glossy appearance of \cref{fig:full_param_wildlight} due to the dominance of the clearcoat and clearcoat-gloss parameters. Details in \cref{sc:BRDF_details}, meshes rendered with \cite{Sketchfab}.}
\label{fig:wildlight_pathology}
\end{figure*}
\begin{figure*}
    \begin{subfigure}[b]{0.30\textwidth}
    \centering
    \includegraphics[width=0.80\textwidth]{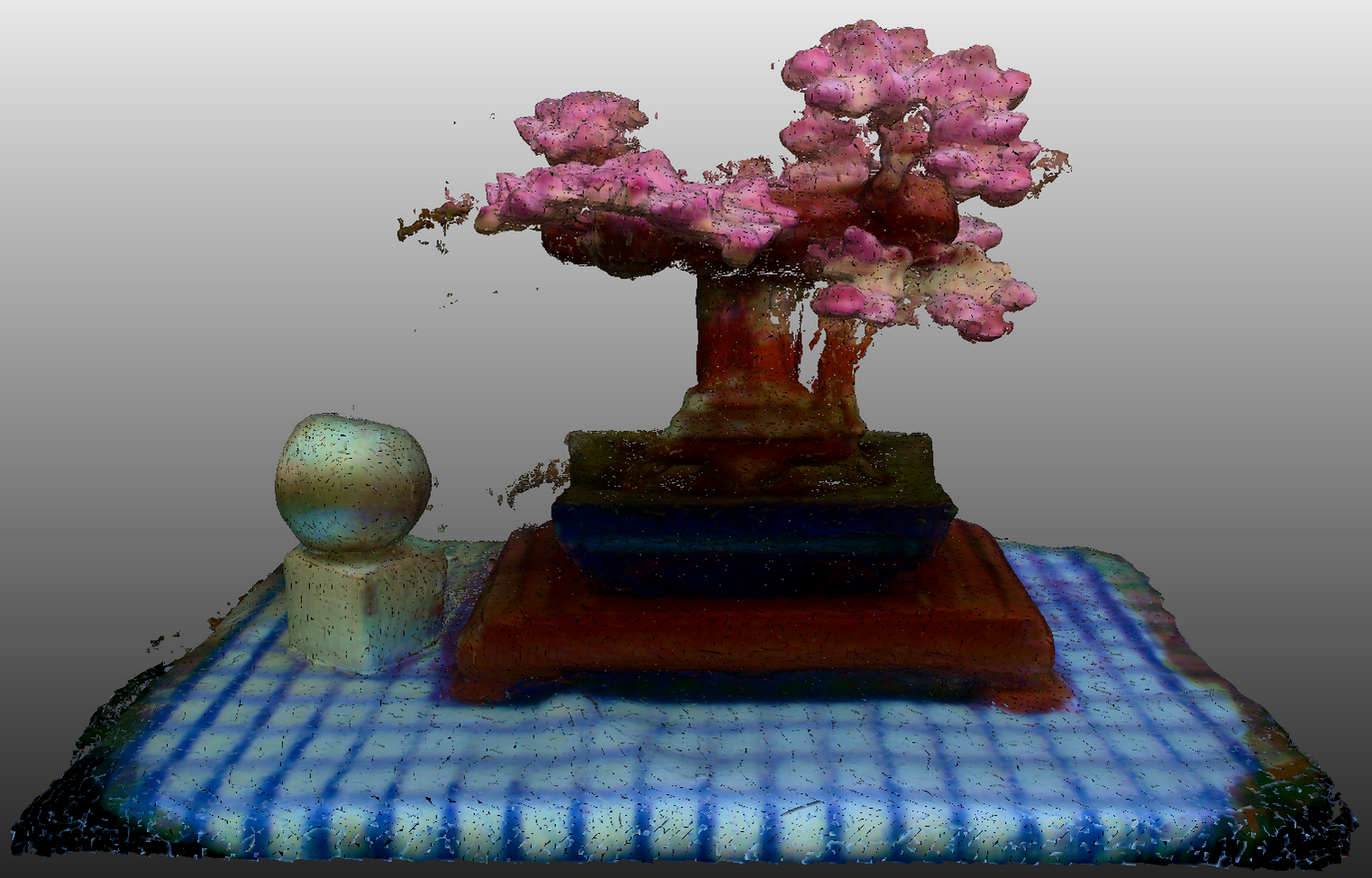}
    \caption{ \adashell}
    \label{fig:volsdf_base_color}
    \end{subfigure}
    \begin{subfigure}[b]{0.30\textwidth}
    \centering
    \includegraphics[width=0.80\textwidth]{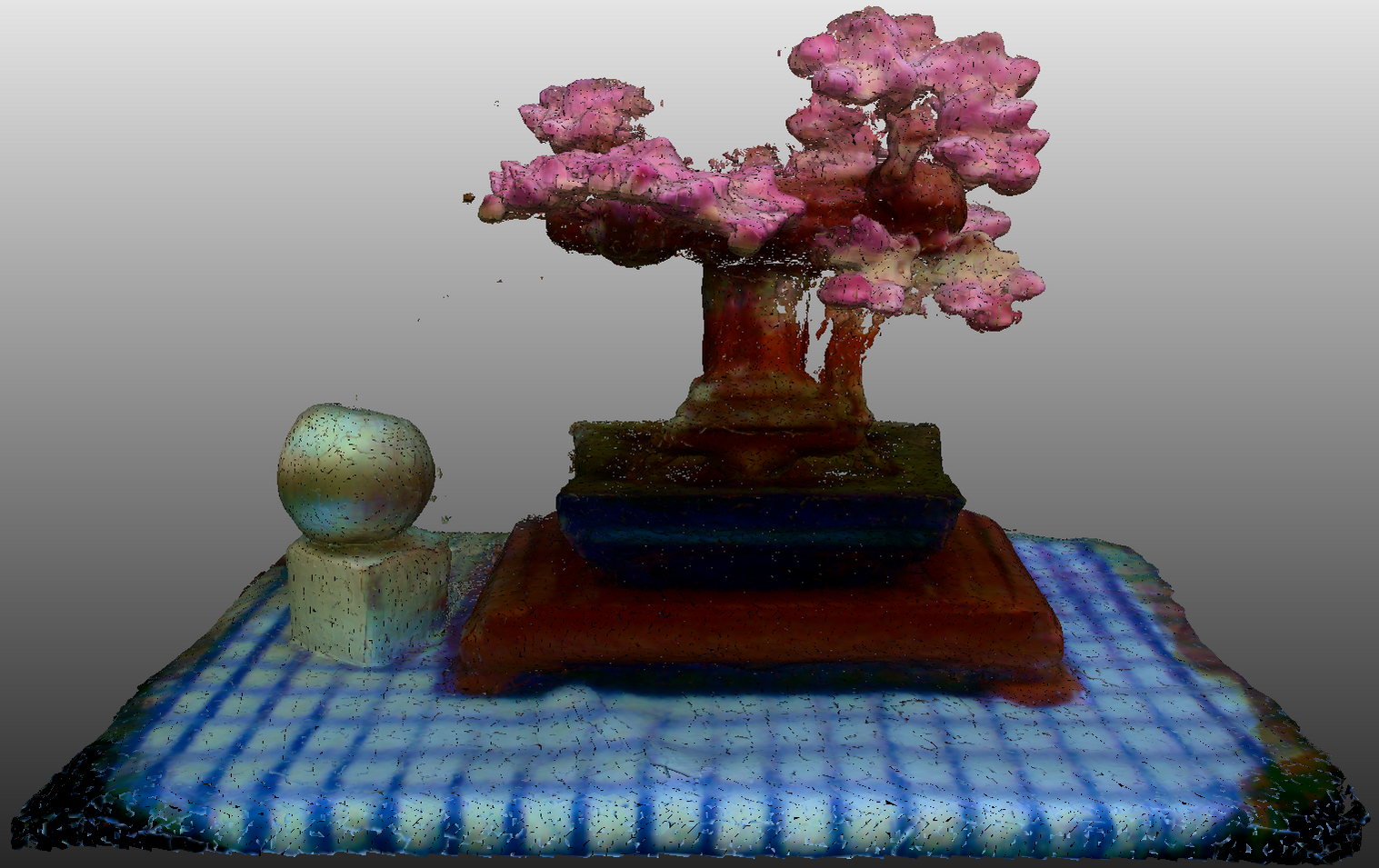}
    \caption{\volsdf}
    \label{fig:adashell_base_color}
    \end{subfigure}
    \begin{subfigure}[b]{0.30\textwidth}
    \centering
    \includegraphics[width=0.80\textwidth]{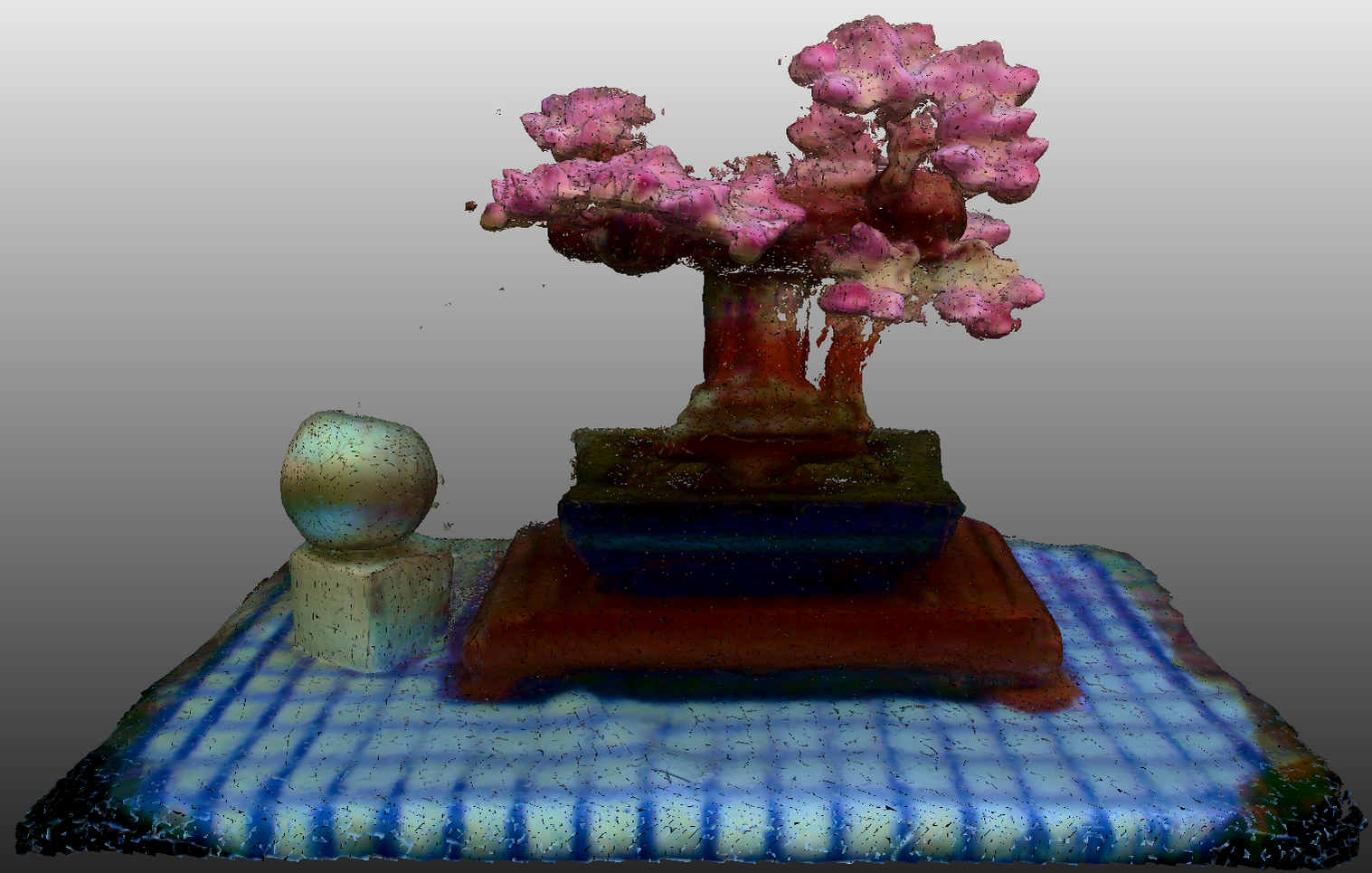}
    \caption{\neus}
    \label{fig:neus_basecolor}
    \end{subfigure}
\caption{\textbf{All the pipelines} can be used to extract the ``base-color'' of the scene. We calculate texture of the meshes from the radiance at convergence (PSNR 27.5+) for one of the scenes in \cref{fig:teaser}. The textured meshes are rendered with MeshLab\cite{meshlab}.}
\label{fig:baking_base_color}
\end{figure*}
Multi-illumination images captured by our camera system can be used to estimate surface reflectance properties. We  recover a truncated Disney BRDF model(\cite{burley2012physically}. Our model consists of a per pixel specular albedo, a diffuse RGB albedo, and a roughness value to interpret the observed appearance under varying illumination. To estimate the spatially varying reflectance, we first train a model (\adashell, \volsdf or \neus ) to convergence to learn the appearance as radiance. At convergence, the first channel $\mathcal{S}$ of the intrinsic network $\mathcal{N}$ encodes the geometry and the appearance network $\mathcal{A}$ encodes the radiance. We use two of the embedding channels of $\mathcal{E}$ to predict the roughness and specular albedo at every point on the scene. The diffuse albedo is obtained as the output of the converged appearance network $\mathcal{A}$. To calculate the appearance, we apply the shading model (\cite{burley2012physically}) to calculate the color at every sample along a ray and volumetrically composite them using \cref{eq:nerf_volumetric_rendering} to infer appearance as reflectance. \Cref{fig:schematic_of_end_to_end} describes our steps graphically.
\newline \indent Optimizing for the full set of the Disney BRDF parameters, following \cite{cheng2023wildlight} did not work with our aproach as the optimization often got stuck at poor local minima. \Cref{fig:full_param_wildlight} shows one instance of optimizing the pipeline of \cite{cheng2023wildlight}, where the strengths of the recovered `clearcoat' and `clearcoat-gloss' parameters dominated over the optimization of the other parameters, resulting in a waxy appearance. Choosing a more conservative set of parameters (only `base-color', `specular' and `roughness') in \cref{fig:reduced_param_wildlight} led to a  more realistic appearance. WildLight\cite{cheng2023wildlight} is based on \cite{wang2021neus} -- we substituted \cite{wang2021neus} with \neus and the apperance model of \cite{cheng2023wildlight} was not changed, minimizing the chances of introduction of a bug causing the artifact. 
\newline \indent Our process of generating texture and material properties roughly follows the methods described by \cite{cheng2023wildlight} and \cite{nerfstudio}. We proceed through the following steps:
\begin{enumerate}
\item At convergence (see \cref{fig:schematic_of_end_to_end}), we extracted the scene geometry using the method described in \cite{Mescheder_2019_CVPR}. 
\item We calculate a depth mask by thresholding the depth images at every training view with an estimate of the scene depth to segregate the foreground from the background.
\item Next, we cull the resulting triangular mesh (step 1) by projecting rays from every unmasked (foreground) pixel corresponding to all the camera views. This lets us extract the main subject of our scene as a mesh. We use Embree\cite{woop2013embree} to implement this. 
\item We generate texture coordinates on the culled mesh using ``Smart UV Unwrap'' function from \cite{blender}. These results were qualitatively better than \cite{xatlas} and our  implementation of \cite{srinivasan2023nuvo}. We then rasterize the culled mesh from step 3 to get points on the surface corresponding to the texture coordinates.
\item We project each of these surface points back on to each of the training views to get the image coordinates. Rays originating from a rasterized surface point and intersecting the surface before reaching the camera are removed to preserve self occlusion.
\item For all the valid projected points, we cast a ray onto the scene and use either \adashell, \volsdf, or \neus to generate the color at the pixel along the ray using \cref{eq:nerf_volumetric_rendering}. This is repeated for all the training views.
\item  At the end of the previous step we have several measurements of colors at every texture coordinate of the scene. We apply a median filter (per color channel) to choose the color -- taking averages or maxima of the samples introduces artifacts. If using the radiance as texture is sufficient (often the case for diffuse scenes) this textured mesh can be exported. \Cref{fig:baking_base_color} demonstrates using each of  \adashell, \volsdf and, \neus to calculate the diffuse color of the scene in \cref{fig:teaser}. 
\item To generate material textures, we follow the same procedures with the corresponding material channels after \adashell, \volsdf or \neus has been trained on multi illumination images using the schedule outlined in \cref{fig:schematic_of_end_to_end}. 
\item The material properties are also volumetrically composited using \cref{eq:nerf_volumetric_rendering} and median filtered like the base colors. This is different from just querying the value of the network at the estimated surface point in \cite{cheng2023wildlight}. 
\end{enumerate}
We use \cite{Sketchfab}, a web browser based tool that supports physically based rendering with the Disney BRDF parameters, and \cite{blender}  to generate the images in \cref{fig:wildlight_pathology,fig:teaser} respectively.  

\section{A multi-flash stereo camera}\label{sc:app_hardware_platform}
We capture the scene using a binocular stereo camera pair with a ring of lights that can be flashed at high intensity. For our prototype, we use a pair of machine vision cameras (\cite{grasshopper3_usb3}) with a 1'', 4MP CMOS imaging sensor of resolution of 2048 $\times$ 2048 pixels. As we focus mainly on small scenes, we use two sets of lenses that yield a narrow field of view -- 12mm and 16mm fixed focal length lens (\cite{edmund_optics}). We use 80W 5600K white LEDs (\cite{cree_LED}) flashed by a high-current DC power supply switched though MOSFETs controlled with an Arduino microcontroller. At each pose of our rig, we captured 12 images with each of the flash lights on (one light at a time) and one HDR image per camera. The cameras are configured to return a 12 bit Bayer image which is then de-Bayered to yield a 16 bit RGB image. 
\newline \indent For the HDR images, we performed a sweep of exposures from the sensor's maximum (22580 microseconds) in 8 stops and used \cite{mertens2007exposure} to fuse the exposures captured with ambient illumination (fluorescent light panels in a room). Following the recommendations of \cite{jensen2014DTU} we used an f-stop of 2.8 to ensure the whole scene is in the depth of field of the sensors. We found the recommendations from \cite{mildenhall2022rawnerf} to be incompatible with our pipeline, so we used Reinhard tone-mapping (\cite{reinhard2002}) to re-interpret the HDR images. Our image localization pipeline, and stereo matching also worked better with tonemapped images.
\newline \indent  We set the left and right cameras to be triggered simultaneously by an external synchronization signal. We configured the camera frame acquisition and the illumination control programs to run in the same thread and synchronized the frame acquisition with the flashes through blocking function calls. \Cref{fig:MFC_schematic} presents a schematic of our prototype device. 
\newline 
\indent Through experiments we observed that the vignetting at the edges of the frames were detrimental to the quality of reconstruction, so we only binned the central 1536 $\times$ 1536 pixels. A 16bit $1536\times1536$ frame saved as a PNG image was often larget than 10MB. To achieve a faster capture and training time without sacrificing the field of view, we down sampled the images to a resolution of $768\times768$ pixels for our experiments. Centered crops of our initial larger frames lead to failures of our pose-estimation pipelines due to the field of view being too narrow(\cref{sc:dataset}), so we chose to down sample the images instead. For the images lit by a single LED, we used the camera's auto exposure function to calculate an admissible exposure for the scene and used 80\% of the calculated exposure time for imaging -- the built-in auto-exposure algorithm tended to over-expose the images a bit. Estimating the exposure takes about 2 seconds. Once the exposure value is calculated, it is used for all of the 12 flashes for each camera.
\newline \indent  Several instances of these RGBD tuples are collected and the colored depth maps are registered in the 3D space in two stages -- first coarsely using FGR~\cite{Zhou2016FGR} and then refined by optimizing a pose graph\cite{Choi2015Posegraph}. At the end of this global registration and odometry step, we retain a reprojection error of about 5 - 10 pixels. If the reprojection errors are not addressed, they will cause the final assets to have smudged color textures.  To address it, we independently align the color images using image-feature based alignment techniques common in multi-view stereo (\cite{sarlin2019hloc, schoenberger2016colmap}), so that a sub 1 pixel mean squared reprojection error is attained. The cameras aligned in the image-space are then robustly transformed to the world space poses using RANSAC\cite{fischler1981RANSAC} with Umeyama-Kabsch's algorithm\cite{umeyama1991}. Finally, we mask out the specular parts of the aligned images and use ColorICP~\cite{park2017coloredICP} to refine the poses. The final refinement step helps remove any  small offset in the camera poses introduced by the robust alignment step. A subset of the data collected can be viewed on the project website.

\subsection{Identifying pixels along depth edges}
To identify pixels along depth edges, we follow \cite{raskar2004MFC} and derive per-pixel likelihoods of depth edges. Assuming that the flashes are point light sources and the scene is Lambertian, we can model the observed image intensity for the $k^{\rm{th}}$ light illuminating a point $\mathbf{x}$ with reflectance $\rho(\mathbf{x})$ on the object as 
\begin{align}\label{eq:image_formation}
    \mathbf{I}_k(\mathbf{x}) = \mu_k \rho(\mathbf{x}) \langle \mathbf{l}_k(\mathbf{x}), \mathbf{n}(\mathbf{x}) \rangle
\end{align}
where $\mu_k$ is the intensity of the $k^{\rm{th}}$ source and $\mathbf{l}_k(\mathbf{x})$ is the normalized light vector at the surface point. $\mathbf{I}_k(\mathbf{x})$ is the image with the ambient component removed. With this, we can calculate a ratio image across all the illumination sources
\begin{align}\label{eq:ratio_images}
    \mathbf{R}(\mathbf{x}) = \dfrac{\mathbf{I}_k(\mathbf{x})}{\mathbf{I}_{max}(\mathbf{x})} = \frac{\mu_k  \langle \mathbf{l}_k(\mathbf{x}), \mathbf{n}(\mathbf{x}) \rangle}{\max_i (\mu_i  \langle \mathbf{l}_i(\mathbf{x}), \mathbf{n}(\mathbf{x}) \rangle)}
\end{align}
It is clear that the ratio image $ \mathbf{R}(\mathbf{x})$ of a surface point is exclusively a function of the local geometry. As the light source to camera baselines are much smaller than the camera to scene distance, except for a few detached shadows and inter-reflections, the ratio images (\cref{eq:ratio_images}) are more sensitive to the variations in geometry than any other parameters. We exploit this effect to look for pixels with largest change in intensity along the direction of the epipolar line between the camera and the light source on the image. This yields a per-light confidence value of whether $\mathbf{x}$ is located on a depth edge or not. Across all 12 illumination sources, we extract the maximum values of the confidences as the depth edge maps. Unlike \cite{raskar2004MFC}, we use 12 illumination sources $30^\circ$ apart, and we do not threshold the confidence values to extract a binary edge map. This lets us extract more edges especially for our narrow depth of field imaging system and gets rid of hyper parameters used for thresholding and connecting the edges.
\newline \indent Often parts of our scene violate the assumption of Lambertian reflectances resulting in spurious depth edges. When we use depth edges for sampling, these errors do not affect the accuracy of our pipeline. When using depth edges for enhancing stereo matching (\cref{sc:mfc_stereo}) we ensure that the stereo pairs do not contain too many of these spurious edge labels to introduce noise in our depth maps. 
\indent \subsection{Identifying patches with non-Lambertian reflectances}
We modified the definition of differential images in the context of near-field photometric stereo introduced by \cite{chandraker2012differential,liu2018near} to identify non-Lambertian patches. Assuming uniform Lambertian reflectances, \cref{eq:image_formation} can be expanded as 
\begin{align}\label{eq:image_formation_expanded}
        \mathbf{I}_k(\mathbf{x}) = \mu_k^* \rho(\mathbf{x})  \mathbf{n}(\mathbf{x})^T \dfrac{\mathbf{s}_k-\mathbf{x}}{|\mathbf{s}_{k}-\mathbf{x}|^3}
\end{align}
where $\mathbf{s}_k$ is the location and $\mu_k^{*}$ is the power of the $k^{\rm{th}}$ light source. We define the differential images as $\mathbf{I}_t = \frac{\partial \mathbf{I}}{\partial \mathbf{s}} \mathbf{s}_t$ where, $\mathbf{s}_t = \frac{\partial \mathbf{s}}{\partial t}$, which when applied to \cref{eq:image_formation_expanded} can be expanded as
\begin{align}\label{eq:liu_diff_images}
    \mathbf{I}_t(\mathbf{x}) = \mathbf{I}(\mathbf{x})\dfrac{\mathbf{n}^T \mathbf{s}_t}{\mathbf{n}^T(\mathbf{s-x})} - 3\mathbf{I}(\mathbf{x})\dfrac{(\mathbf{s-x})^T\mathbf{s}_t}{|\mathbf{s-x}|^2}
\end{align}
Observing that the light sources move in a circle around the center of projection on the imaging plane, $\mathbf{s}^T\mathbf{s}_t = 0$. Also, the second term of \cref{eq:liu_diff_images} is exceedingly small given that the plane spanned by $\mathbf{s}_t$ is parallel to the imaging plane and our choice of lenses limit the field of view of the cameras. The second term is further attenuated by the denominator $|\mathbf{s-x}|^2$ because the camera-to-light baselines ($\mathbf{s}$) are at least an order of magnitude smaller than the camera to object distance ($\mathbf{x}$). As a result, under isotropic reflectances (Lambertian assumed for this analysis) the differential images $ \mathbf{I}_t(\mathbf{x})$ are invariant to circular light motions. Any observed variance therefore can be attributed to the violations of our isotropic BRDF assumptions. We identify specular patches by measuring the variance of this quantity across the 12 instances of the flashlit images.   
\newline \indent Although our pipelines for identifying depth edges and patches of varying appearances demonstrate satisfactory qualitative performance, sometimes they yield wrong labels because \cref{eq:ratio_images,eq:liu_diff_images} do not include additional terms for spatially varying BRDFs and interreflections respectively. These errors do not have any significant effect in our reconstruction pipeline as we use this information to generate samples during different phases of training to minimize photometric losses and we do not directly infer shape or reflectances from these steps. 
\subsection{Difference between \cite{feris2005discontinuity,raskar2004MFC} and our hardware}
\textbf{\cite{raskar2004MFC}} was the first to propose pairing flashes with cameras and laid the groundwork for identifying depth edges from multi-flash images from a single viewpoint. However, \cite{raskar2004MFC} considered a monocular camera and only four flashes along the horizontal and vertical directions of the camera in the demonstrated device. Researchers (see e.g. \cite{chaudhury2024shape}) have since extended it by placing multiple light sources far apart from a monocular camera and have demonstrated locating depth edges on objects with strictly Lambertian reflectances. In this work, we retain the original light and camera configuration from \cite{raskar2004MFC} and increase the number of lights from four to 12. 
\newline \indent \textbf{\cite{feris2005discontinuity}} also investigated a stereo camera in a multi-flash configuration aimed at edge preserving stereo depth maps. They do not extend the application to synthesizing geometry or appearance by capturing and assimilating multiple views of the scene. For obtaining stereo depth maps, we use \cite{xu2022gmflow}, which performs much better than conventional stereo matching (\cite{hirschmuller2005accurate,zabih1994non}) largely deployed in off-the shelf systems (\cite{Keselman2017_realsense}). 
\newline \indent \textbf{Both \cite{raskar2004MFC,feris2004specular}} discuss methods to detect specularities (termed ``material edges'') through different transforms of the multi-light images. However, we achieve a more continuous circular motion of the lights around the cameras, so we choose to use the photometric invariants described by \cite{chandraker2012differential} instead. 
%
\end{document}